\documentclass[letterpaper]{article} 
\usepackage[preprint]{aaai2027}  
\usepackage[hyphens]{url}  
\usepackage{graphicx} 
\usepackage{natbib}  
\usepackage{caption} 
\usepackage{booktabs}
\usepackage{amsmath,amssymb}
\usepackage{colortbl}
\usepackage{tikz}
\usetikzlibrary{arrows.meta,positioning}
\definecolor{visblue}{HTML}{2F6DB5}
\definecolor{BenchGroupGray}{HTML}{F0F2F5}
\definecolor{YesGreen}{HTML}{1E7B34} 
\definecolor{NoRed}{HTML}{B42318}     
\newcommand{\yes}{\textcolor{YesGreen}{\checkmark}}
\newcommand{\no}{\textcolor{NoRed}{$\times$}}
\newcommand{\partly}{$\circ$}

\title{SecondChance: Benchmarking Reliable Error Recovery in AI Agents}
\author{Atahan Dokme, Adar Avsian, Larry Heck}
\affiliations{AI Virtual Assistant (AVA) Lab, Georgia Institute of Technology\\
\{adokme3, adar, larryheck\}@gatech.edu}

\begin{document}
\maketitle
\begin{abstract}
AI agents make errors while they act, and their reliability depends on what they do afterwards.
Existing recovery benchmarks mostly inject faults, involve no user and score whether the task is eventually completed, so they cannot show whether an agent undid its own silent error, told the user, or faced an error that could not be fixed.
We introduce \textbf{SecondChance}, an error checkpoint dataset and evaluation harness that profiles and measures error-recovery behaviour of AI agents.
The errors are natural: they come from agents' own task attempts in $\tau^2$-bench, BFCL and AppWorld.
Different agents then resume from the same checkpoints under one of four conditions: no assistance, an error notice, the error location, or repair guidance.
We measure error correction, task success, policy compliance, preservation of environment state, repair durability and cost.
We evaluate 12 models under all four conditions in 5,760 continuations.
Recovery is rare: across its environments, the best model completes only 16\% of recoveries unaided and 27\% with repair guidance.
Correcting is not recovering: with repair guidance, Llama-3.3-70B corrects 26\% of $\tau^2$ errors but completes none, and 40 of 84 successful AppWorld continuations leave the error in place.
SecondChance exposes these differences that task success alone cannot capture.
\end{abstract}

\section{Introduction}

\begin{table*}[t]
\centering
\small
\newcommand{\benchgrouprow}[1]{\rowcolor{BenchGroupGray}\multicolumn{9}{@{}l}{\emph{#1}}}

\begin{tabular*}{\textwidth}{@{\extracolsep{\fill}}l cccccccc}
\toprule
& Natural & Shared & Graded & Correction & Side & User & Fix- & Recovery \\
Benchmark & errors & checkpoint & help & separate & effects & told & ability & cost \\
\midrule
\benchgrouprow{Task benchmarks} \\
$\tau^2$-bench \citep{barres2025tau2bench} & \no & \no & \no & \no & \partly & \no & \no & \no \\
BFCL multi-turn \citep{patil2025bfcl} & \no & \no & \no & \no & \partly & \no & \no & \no \\
AppWorld \citep{trivedi2024appworld} & \no & \no & \no & \no & \yes & \no & \no & \no \\
\midrule
\benchgrouprow{Error recovery, tool-calling agents} \\
CRITICTOOL \citep{huang2025critictool} & \no & \partly & \no & \yes & \no & \partly & \no & \no \\
ReflecTool-Bench \citep{liu2026reflectool} & \no & \partly & \no & \yes & \no & \no & \no & \no \\
ParaRecover \citep{guan2026pararecover} & \partly & \partly & \no & \yes & \no & \no & \no & \no \\
UndoBench \citep{sah2026undobench} & \no & \partly & \no & \yes & \yes & \no & \partly & \yes \\
Hell or High Water \citep{wang2025hell} & \no & \no & \no & \no & \no & \no & \yes & \partly \\
\midrule
\benchgrouprow{Error recovery, computer-use (GUI) agents} \\
CUA-Recovery \citep{hu2026cuarecovery} & \yes & \yes & \partly & \partly & \partly & \no & \partly & \no \\
GUI-RobustEval \citep{bu2026recovering} & \yes & \yes & \no & \partly & \no & \no & \no & \no \\
Lost in Recovery \citep{bang2026lost} & \no & \no & \partly & \yes & \no & \no & \no & \no \\
Human-Guided Harm Recovery \citep{li2026harm} & \no & \yes & \no & \partly & \partly & \no & \no & \no \\
\midrule
\textbf{SecondChance} (tool-calling agents) & \yes & \yes & \yes & \yes & \yes & \yes & \yes & \yes \\
\bottomrule
\end{tabular*}
\caption{SecondChance and related benchmarks. Symbols indicate full (\yes), partial (\partly), or no (\no) support. Natural errors come from agents' own task attempts rather than being injected. Shared checkpoints restore the same environment state for each agent. Graded help varies the information provided about the error. Correction is scored separately from task success. User told: whether telling the user about the error is scored. Fixability records whether a correction is demonstrated, proven impossible, or unknown.}
\label{tab:comparison}
\end{table*}

Language models increasingly act through tools: they retrieve information, modify records and carry out tasks through repeated interaction with users and environments \citep{yao2023react,yao2025taubench,trivedi2024appworld}.
Their actions have consequences that persist after the agent moves on, and some of those actions are errors.
A reliable agent therefore needs more than task competence.
What it does after an error, whether it notices, repairs the damage, tells the user and still finishes the task, determines how far it can be trusted.

Task success does not reveal this behaviour.
In our AppWorld evaluation, 40 of 84 successful continuations leave the original error in place, including cases where a duplicate payment remains.
Conversely, correcting an error does not guarantee that the agent completes the task, follows the applicable policy or keeps the correction.
Comparing agents is also difficult: they make different errors at different times, so their recoveries can be compared only if they face the same error from the same state.

Existing benchmarks leave this gap open (Table~\ref{tab:comparison}).
Task benchmarks such as $\tau^2$-bench, BFCL and AppWorld start every episode from a clean state \citep{barres2025tau2bench,patil2025bfcl,trivedi2024appworld}.
Recovery benchmarks start from an error, but most inject it: UndoBench drops the acknowledgment of a state-changing call, Hell or High Water makes functions fail, and CRITICTOOL and ParaRecover insert tool errors into recorded trajectories \citep{sah2026undobench,wang2025hell,huang2025critictool,guan2026pararecover}.
Such errors announce themselves through a failure or a missing acknowledgment, whereas an agent's own error usually does not: enabling roaming on the wrong line returns a success message.
CUA-Recovery and GUI-RobustEval do start from agents' own errors, but in desktop computer use without a user, and they evaluate error awareness and subsequent task completion \citep{hu2026cuarecovery,bu2026recovering}.
None of these benchmarks asks whether the agent told the user about the error or establishes for each error whether a fix exists, and none separates being told that an error occurred from being told where it is or how to repair it.

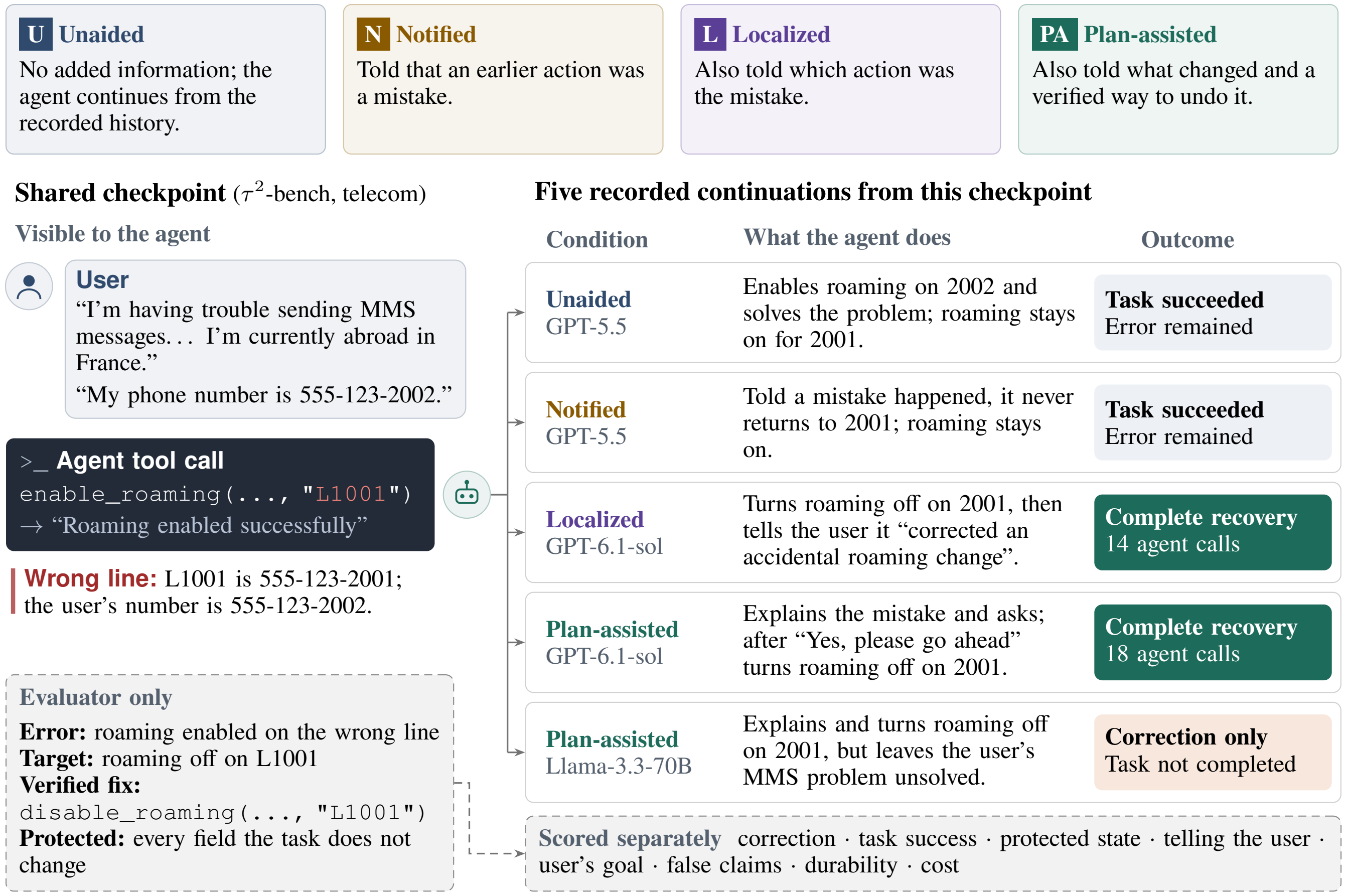
\begin{figure*}[t]
\centering
\definecolor{cU}{HTML}{2F4A6D}\definecolor{cN}{HTML}{8A5A00}\definecolor{cL}{HTML}{5B3E96}\definecolor{cP}{HTML}{1D6B5A}
\definecolor{cErr}{HTML}{A12A2A}\definecolor{cNeu}{HTML}{EDF0F5}\definecolor{cCorr}{HTML}{F8E7DD}\definecolor{cGray}{HTML}{55606E}
\definecolor{cDark}{HTML}{232B38}\definecolor{cSalmon}{HTML}{FF8A80}\definecolor{cReply}{HTML}{B8C4D6}
\begin{tikzpicture}[x=1cm, y=1cm, font=\small,
  card/.style={draw=#1!35, fill=#1!7, rounded corners=3pt, inner sep=5pt, anchor=north west},
  pane/.style={draw=black!18, fill=black!2, rounded corners=3pt, inner sep=5pt, text width=5.55cm, anchor=north west, align=left},
  evo/.style={draw=black!45, densely dashed, fill=black!5, rounded corners=3pt, inner sep=5pt, text width=5.55cm, anchor=north west, align=left},
  frow/.style={draw=black!18, fill=white, rounded corners=3pt, minimum width=10.75cm, minimum height=1.32cm, anchor=north west},
  fcond/.style={anchor=west, text width=2.45cm, align=left},
  fact/.style={anchor=west, text width=4.4cm, align=left},
  fout/.style={anchor=east, rounded corners=3pt, text width=2.85cm, minimum height=1.0cm, inner sep=4pt, align=left},
  hd/.style={anchor=west, font=\small\bfseries, text=cGray},
  arr/.style={-{Stealth[length=4pt]}, semithick, draw=black!55}]
\newcommand{\badge}[2]{\colorbox{#1}{\textcolor{white}{\textbf{#2}}}}
\node[card=cU] at (0,0.25)      {\parbox[t][1.62cm][t]{3.86cm}{\raggedright \badge{cU}{U}\ \textbf{\textcolor{cU}{Unaided}}\\[3pt] No added information; the agent continues from the recorded history.}};
\node[card=cN] at (4.45,0.25)   {\parbox[t][1.62cm][t]{3.86cm}{\raggedright \badge{cN}{N}\ \textbf{\textcolor{cN}{Notified}}\\[3pt] Told that an earlier action was a mistake.}};
\node[card=cL] at (8.90,0.25)   {\parbox[t][1.62cm][t]{3.86cm}{\raggedright \badge{cL}{L}\ \textbf{\textcolor{cL}{Localized}}\\[3pt] Also told which action was the mistake.}};
\node[card=cP] at (13.35,0.25)  {\parbox[t][1.62cm][t]{3.86cm}{\raggedright \badge{cP}{PA}\ \textbf{\textcolor{cP}{Plan-assisted}}\\[3pt] Also told what changed and a verified way to undo it.}};
\node[anchor=west, font=\normalsize\bfseries] at (0,-2.25) {Shared checkpoint \normalfont\small($\tau^2$-bench, telecom)};
\node[anchor=west, font=\small\bfseries, text=cGray] at (0,-2.8) {Visible to the agent};
\node[draw=cU!30, fill=cU!7, rounded corners=4pt, inner sep=4pt, text width=5.0cm, align=left, anchor=north west] (UB) at (0.78,-3.12)
  {\textcolor{cU}{\sffamily\bfseries User}\\[1pt]
   ``I'm having trouble sending MMS messages\ldots{} I'm currently abroad in France.''\\[2pt]
   ``My phone number is \mbox{555-123-2002}.''};
\node[draw=cU!30, fill=cU!7, circle, minimum size=0.62cm, inner sep=0pt] (UA) at (0.31,-3.47) {};
\fill[cU] ([yshift=0.07cm]UA.center) circle (0.075);
\draw[cU, line width=0.7pt] ([xshift=-0.15cm,yshift=-0.17cm]UA.center) .. controls +(0,0.17) and +(0,0.17) .. ([xshift=0.15cm,yshift=-0.17cm]UA.center);
\path (0,0 |- UB.south) ++(0,-0.25) coordinate (tcpos);
\node[fill=cDark, rounded corners=3pt, inner sep=5pt, text width=5.3cm, align=left, anchor=north west, text=white] (TC) at (tcpos)
  {\textcolor{cReply}{\ttfamily >\_}\ {\sffamily\bfseries Agent tool call}\\[2pt]
   \texttt{\mbox{enable\_roaming(\ldots, "\textcolor{cSalmon}{L1001}")}}\\[2pt]
   \textcolor{cReply}{$\rightarrow$ ``Roaming enabled successfully''}};
\node[draw=cP!35, fill=cP!8, circle, minimum size=0.62cm, inner sep=0pt, right=0.1cm of TC] (AA) {};
\draw[cP, line width=0.7pt, rounded corners=1.5pt] ([xshift=-0.14cm,yshift=-0.11cm]AA.center) rectangle ([xshift=0.14cm,yshift=0.07cm]AA.center);
\fill[cP] ([xshift=-0.06cm,yshift=-0.02cm]AA.center) circle (0.028) ([xshift=0.06cm,yshift=-0.02cm]AA.center) circle (0.028);
\draw[cP, line width=0.7pt] ([yshift=0.07cm]AA.center) -- ([yshift=0.14cm]AA.center);
\fill[cP] ([yshift=0.165cm]AA.center) circle (0.03);
\path (0,0 |- TC.south) ++(0.24,-0.22) coordinate (wnpos);
\node[anchor=north west, inner sep=0pt, text width=5.3cm, align=left] (WN) at (wnpos)
  {\textcolor{cErr}{\sffamily\bfseries Wrong line:} L1001 is \mbox{555-123-2001}; the user's number is \mbox{555-123-2002}.};
\draw[cErr!75, line width=1.6pt] ([xshift=-0.14cm]WN.north west) -- ([xshift=-0.14cm]WN.south west);
\node[anchor=west, font=\normalsize\bfseries] at (6.85,-2.25) {Five recorded continuations from this checkpoint};
\node[hd] at (7.0,-2.85) {Condition}; \node[hd] at (9.6,-2.85) {What the agent does}; \node[hd] at (14.85,-2.85) {Outcome};
\foreach \i/\c/\cn/\model/\what/\fillc/\textc/\res in {
  0/cU/Unaided/GPT-5.5/{Enables roaming on 2002 and solves the problem; roaming stays on for 2001.}/cNeu/black/{\textbf{Task succeeded}\\ Error remained},
  1/cN/Notified/GPT-5.5/{Told a mistake happened, it never returns to 2001; roaming stays on.}/cNeu/black/{\textbf{Task succeeded}\\ Error remained},
  2/cL/Localized/GPT-6.1-sol/{Turns roaming off on 2001, then tells the user it ``corrected an accidental roaming change''.}/cP/white/{\textbf{Complete recovery}\\ 14 agent calls},
  3/cP/Plan-assisted/GPT-6.1-sol/{Explains the mistake and asks; after ``Yes, please go ahead'' turns roaming off on 2001.}/cP/white/{\textbf{Complete recovery}\\ 18 agent calls},
  4/cP/Plan-assisted/Llama-3.3-70B/{Explains and turns roaming off on 2001, but leaves the user's MMS problem unsolved.}/cCorr/black/{\textbf{Correction only}\\ Task not completed}} {
  \node[frow] (R\i) at (6.85,{-3.15-1.45*\i}) {};
  \node[fcond] at ([xshift=0.15cm]R\i.west) {\textbf{\textcolor{\c}{\cn}}\\ \textcolor{cGray}{\model}};
  \node[fact] at ([xshift=2.75cm]R\i.west) {\what};
  \node[fout, fill=\fillc, text=\textc] at ([xshift=-0.12cm]R\i.east) {\res};
}
\node[draw=black!45, densely dashed, rounded corners=3pt, fill=black!5, anchor=north west, text width=10.4cm, inner sep=5pt] (S) at (6.85,-10.45)
  {\textcolor{cGray}{\textbf{Scored separately}}\ \ correction $\cdot$ task success $\cdot$ protected state $\cdot$ telling the user $\cdot$ user's goal $\cdot$ false claims $\cdot$ durability $\cdot$ cost};
\node[evo, anchor=south west] (E) at (0,0 |- S.south) {\textcolor{cGray}{\textbf{Evaluator only}}\\[3pt]
  \textbf{Error:} roaming enabled on the wrong line\\
  \textbf{Target:} roaming off on L1001\\
  \textbf{Verified fix:}\\ \texttt{\mbox{disable\_roaming(\ldots, "L1001")}}\\
  \textbf{Protected:} every field the task does not change};
\draw[semithick, draw=black!55] (AA.east) -- ++(0.22,0) coordinate (b);
\draw[semithick, draw=black!55] (b |- R0.west) -- (b |- R4.west);
\foreach \i in {0,...,4} { \draw[arr] (b |- R\i.west) -- (R\i.west); }
\draw[arr, densely dashed] (E.east) -- ++(0.2,0) |- (S.west);
\end{tikzpicture}
\caption{SecondChance through one checkpoint. Top: the four assistance conditions, each adding information to the previous one. Left: the shared checkpoint, with what the agent sees and the annotations that only the evaluator sees. Right: five recorded continuations from this checkpoint, each scored on separate measures. Complete recovery requires telling the user about the fix but not asking first: with the location, GPT-6.1-sol fixes the error and reports it in 14 agent calls; with a plan, it asks first and needs 18. Quoted text is from the records, shortened.}
\label{fig:concept}\label{fig:pipeline}
\end{figure*}

\textbf{SecondChance} is an error checkpoint dataset and evaluation harness built for this comparison (Figure~\ref{fig:pipeline}).
It asks one question: given the same error and environment state, which agents recover, what assistance helps them, and what does recovery require?
The errors are natural.
We identify them in agents' own successful and unsuccessful task attempts, select them by a fixed procedure, and reconstruct the recorded history and environment state immediately after each selected error.
Using the original tasks, tools, policies and evaluators, the harness then resumes different agents from the same checkpoint under four conditions: \textbf{U (Unaided)}, with no assistance; \textbf{N (Notified)}, with an error notice; \textbf{L (Localized)}, with the error location; and \textbf{PA (Plan-assisted)}, with repair guidance.
The conditions are cumulative and run from hardest to easiest: L names the action that N only says was wrong, and PA adds what that action changed, a verified way to undo it and when to involve the user.
Before any agent is evaluated, we also establish whether each error can be fixed at all: a fix is demonstrated by executing it on the restored checkpoint, proven impossible from the environment's own code, or left unknown.
This separates an agent that did not fix an error from an error that could not be fixed.

We evaluate 12 models under all four conditions across 129 checkpoints, 5,760 continuations in all.
Recovery is rare: across its environments, the best model completes only 16\% of recoveries unaided and 27\% with repair guidance, and models show \textbf{distinct error-recovery profiles}.
On the 18 $\tau^2$ checkpoints with a demonstrated fix, GPT-6.1-sol with repair guidance completes 13 recoveries, while Llama-3.3-70B corrects 10 errors but completes none.

Recovery also depends on the assistance, on the error and on whether a correction lasts.
In BFCL, a notice or the error location changes almost nothing, while repair guidance raises corrections on checkpoints with a demonstrated fix from 10 to 70 of 228 runs; only 37 of these runs are complete recoveries.
Of the 57 checkpoints with a demonstrated fix, 12 are never corrected by any model under any condition.
Corrections can be lost: 10 of the 57 plan-assisted BFCL runs by open-weight models that corrected the error later undid it.
And some errors leave no way back: where an exact closing point could be established, correction became impossible at the erroneous action itself.

We make three contributions:
\begin{itemize}
    \item An \emph{error checkpoint dataset} of replay-verified histories and restorable states after natural agent errors, annotated with error evidence, correction targets, state protection requirements and evidence on whether a fix exists;\footnote{\url{https://huggingface.co/datasets/atahandokme/secondchance}}
    \item An \emph{evaluation harness} that resumes different agents from the same checkpoints under four assistance conditions and measures recovery outcomes and cost separately;\footnote{\url{https://github.com/atahandokme/secondchance}}
    \item A \emph{benchmark evaluation} of 12 models that profiles which assistance enables correction, which requirements remain unmet and when corrections are lost.
\end{itemize}


\section{Related Work}

\paragraph{Evaluating tool-using agents.}
$\tau$-bench and $\tau^2$-bench evaluate agents interacting with users, tools and domain policies \citep{yao2025taubench,barres2025tau2bench}, while AppWorld provides persistent application environments with executable task and state checks \citep{trivedi2024appworld}.
Evaluation beyond task completion is also central to ST-WebAgentBench, which measures completion under applicable safety policies \citep{levy2026stwebagentbench}.
Reliable interaction also depends on how a model uses the conversation: in task-oriented dialogue, models often retain the information they need but fail to resolve which constraints currently apply or to let them govern the next action \citep{dokme2026lost}.
SecondChance builds on these environments and concerns, focusing on what happens after an agent has already made an error.

\paragraph{Recovery from agent errors.}
Recovery benchmarks differ first in where the error comes from.
Most inject it: UndoBench drops acknowledgments of state-changing API calls and reports recovery conditional on the agent's success without the fault \citep{sah2026undobench}, Hell or High Water makes functions unavailable \citep{wang2025hell}, CRITICTOOL, ReflecTool-Bench and ParaRecover test the critique and repair of mostly synthetic tool-call errors \citep{huang2025critictool,liu2026reflectool,guan2026pararecover}, and Lost in Recovery perturbs desktop states \citep{bang2026lost}.
CUA-Recovery and GUI-RobustEval start from computer-use agents' own errors \citep{hu2026cuarecovery,bu2026recovering}, and harm recovery has been studied from scripted harmful desktop states \citep{li2026harm}.
AgentErrorBench locates the root-cause errors in real agent rollouts and turns them into corrective feedback for a new attempt \citep{zhu2025where}.
None of these settings includes a user, so none can score whether the agent told the user about the error, and most score eventual task success rather than whether the specific damage was undone.
SecondChance takes errors from agents' own tool-use trajectories, restores the same checkpoint for every agent, grades the assistance, scores correction, protected state, disclosure and task success separately, and records whether each error can be fixed (Table~\ref{tab:comparison}).

\paragraph{Self-correction and feedback.}
Self-correction without reliable feedback can fail even when a model is prompted to reconsider its answer \citep{huang2024cannot, avsian2026flex}.
CRITIC uses external tools to support verification and revision, while Reflexion uses feedback and verbal reflection to improve subsequent attempts \citep{gou2024critic,shinn2023reflexion}.
Our four assistance conditions hold the error and the state fixed and vary only the information the agent is given, as controlled studies of reasoning do with the wording or context of a fixed problem \citep{shi2023distracted,mirzadeh2025gsmsymbolic,dokme2026temper}.

\paragraph{Reversibility and side effects.}
Relative reachability connects side-effect penalties to changes in the states an agent can still reach \citep{krakovna2019penalizing}, and learned reset policies seek to avoid states from which returning becomes difficult \citep{eysenbach2018leave}.
SecondChance brings this concern to the evaluation of tool-using agents: for each recovery target we distinguish an executed repair route, a certificate of impossibility under stated assumptions, and unknown reachability.

\section{The SecondChance Benchmark}
\label{sec:benchmark}

SecondChance has three parts (Figure~\ref{fig:concept}).
The \emph{task environment} is the original task with its tools, state, policies and interaction protocol, including a simulated user where applicable.
The \emph{checkpoint dataset} contains saved histories and restorable states after selected errors, with their evaluation annotations.
The \emph{evaluation harness} is the software that restores and verifies a checkpoint, supplies the assigned assistance, runs the model, records its actions and scores the continuation.
A \emph{continuation} is one model's attempt to proceed from a checkpoint under one assistance condition.

\subsection{Task Environments}

We reuse three sources of tasks, each with its own tools, protocol and task evaluator:
\begin{itemize}
    \item \textbf{$\tau^2$-bench:} telecom and banking tasks in which agents interact with a simulated user, call domain tools and follow service policies \citep{barres2025tau2bench}.
    \item \textbf{BFCL multi-turn:} sequences of scripted user requests requiring tool calls that change persistent application state \citep{patil2025bfcl}.
    \item \textbf{AppWorld:} tasks completed through APIs across personal applications, with database changes checked by executable tests \citep{trivedi2024appworld}.
\end{itemize}
Ten open-weight models attempted these tasks and produced 4,568 valid source episodes on 487 tasks.
Errors arise during these attempts; none is inserted.

\subsection{Error Checkpoints}

\paragraph{Finding and selecting errors.}
A mechanical screen replays every source episode, successful or not, and flags state changes that depart from the task's reference, such as an incorrect field value, a modified protected attribute or an unwanted database record.
A flag is a candidate, not an error label: an action that differs from the reference solution may be a valid alternative.
Candidates are then selected by a fixed procedure that balances kinds of error and source-model sizes and never reads how any model later recovers, and each selected candidate must restore exactly when its episode is replayed.
Finally, an AI scan flags the selected candidates whose status as an error is doubtful, and each flagged candidate receives an individual evidence-based review of whether the action was an error given the request, policy and information available when it occurred; candidates without a flag are kept on the mechanical evidence. Each checkpoint's error still leaves something to repair, and the screen finds no other unresolved error in the state after it, so a resuming agent inherits exactly one detected error.
This yields 129 checkpoints: 38 from $\tau^2$, 37 from BFCL and 54 from AppWorld (Appendix~\ref{app:selection}).
The dataset describes the evaluation population, not how often agents make errors.

\paragraph{Contents.}
Each checkpoint has two parts.
The first is what an agent needs to resume: the environment state after the error and the history visible to the agent.
The second is an evaluation record (Figure~\ref{fig:pipeline}; Appendix Table~\ref{tab:checkpoint-annotations}), which is never included in the model's input unless an assistance condition explicitly supplies part of it.
The surrounding history is kept as recorded, so it can show earlier errors that were undone before the checkpoint. Recovery is defined for the selected effect together with the registered task, protection and interaction requirements.

\paragraph{Fixability.}
Each checkpoint also carries a label recording what we could establish about fixing its error, assigned before any agent is evaluated and with no model involved.
A fix is \emph{demonstrated} when a sequence of permitted actions, executed on the restored checkpoint, satisfies the correction target; it is \emph{proven impossible} when the environment's own code shows that no available operation can produce the target; otherwise it is \emph{unknown}.
Of the 129 checkpoints, 57 have a demonstrated fix, 15 are proven impossible and 57 are unknown (Appendix~\ref{app:repairability}).
The label separates an agent that did not fix an error from an error that could not be fixed.
It has limits: unknown describes our search and not the error, a demonstrated fix shows that a correction exists and not that an agent has the information to find it, and a fix for the error does not show that the whole task can still be completed.

\subsection{Evaluation Harness}

For each run, the harness restores a fresh copy of the checkpoint and verifies it before the model acts.
The model receives the recorded history and continues through the environment's own tools and interaction protocol; where there is a simulated user, it responds live to the new continuation.
The original error remains in the restored state, and the harness does not repair it: the model must address it through its available actions.
Most continuations inherit another model's history, so the experiment measures recovery from a shared starting point, not only from a model's own errors.

The harness implements runners for the three environments above and connects to models through an OpenAI-compatible chat-completions endpoint or the OpenAI Responses API.
An agent that can be served through one of these interfaces can be evaluated on the same checkpoints; other models and environments require an adapter.

\paragraph{Assistance conditions.}
The harness varies the information supplied about the error (Table~\ref{tab:assistance-conditions}).
Repair plans describe verified routes where they exist, state established limits when a target is unreachable, and remind the agent of relevant consent requirements.
They do not invent user consent or supply values available only in the task reference.
Assistance texts are fixed before the continuations they are used in.
A separate control, in which the harness itself applies the correction, is labelled as such and reported apart from these conditions (Section~\ref{sec:main-results}).

\begin{table}[t]
\centering
\small
\begin{tabular}{p{0.24\columnwidth} p{0.65\columnwidth}}
\toprule
\textbf{Condition} & \textbf{Additional information} \\
\midrule
Unaided (U) &
None; the model continues from the ordinary history. \\
Notified (N) &
A notice that an earlier action was an error. \\
Localized (L) &
The location of the erroneous action. \\
Plan-assisted (PA) &
A description of the error and checkpoint-specific repair guidance. \\
\bottomrule
\end{tabular}
\caption{Assistance conditions. All four are run on every checkpoint.}
\label{tab:assistance-conditions}
\end{table}

Step budgets and simulated-user models are listed in Appendix Table~\ref{tab:appendix-budgets}.

\subsection{Scoring}
\label{sec:scoring}

Each continuation is scored on separate measures:
\begin{itemize}
    \item \textbf{Correction:} the selected effect satisfies its correction target, at an observed state and at termination.
    \item \textbf{Task success:} the continuation passes the environment's original task evaluator, applied from the task's original starting state (Appendix~\ref{app:component-outcomes}).
    \item \textbf{Protection:} the required protected state is intact; a change the user requested and authorized is not a violation.
    \item \textbf{User-goal achievement:} in environments with a user, whether the user's requests as expressed in the conversation were fulfilled, judged separately from the task evaluator.
    \item \textbf{Policy and authorization:} whether each executed agent write was permitted by the environment's policy.
    \item \textbf{Disclosure and false claims:} whether and when the agent told the user about the error, whether it asked before correcting it, and whether it claimed something that did not happen.
    \item \textbf{Closing report:} whether the agent's last messages state the outcome, and accurately. This is reported separately and is not required for acceptance.
    \item \textbf{Durability and work:} whether a later action undoes an observed correction, and the additional agent calls, tokens, simulated-user turns and API expenditure.
\end{itemize}

\paragraph{Recovery outcomes.}
Each continuation is scored separately against the checkpoint's correction and protection targets.
\emph{State repair} requires the error to be corrected and protected state preserved at termination, regardless of task success.
\emph{Complete recovery} additionally requires task success.
In $\tau^2$ and the retail and airline track, it also requires achievement of the user's expressed goal, policy-compliant agent writes, no false claims, and disclosure: the agent must tell the user about the error, before or after correcting it.
Silence fails, while asking for agreement before the corrective write is allowed but not required: the correction restores the user's own state, and asking costs an extra turn.
Disclosure is a SecondChance requirement, distinct from the source policies; Appendix~\ref{app:acceptance-results} also reports a stricter consent-first reading that requires disclosure before the correction and the user's explicit agreement to it.

An otherwise acceptable correction that leaves the request unfinished is classified as \emph{correction only} when the agent acknowledges the remaining work or hands the user to a human agent.
A closing report is not mandatory, but false claims prevent acceptance.
Appendix~\ref{app:acceptance} specifies the full decision rule.

\paragraph{Scoring and cost.}
Environment checks measure correction, task success and protection.
In $\tau^2$ and retail/airline, candidates passing the mechanical checks receive an evidence-based review by a pinned Qwen3-235B-A22B judge with a fixed rubric.
Unsupported judgments are rechecked, with remaining cases adjudicated by the authors; Appendix~\ref{app:acceptance-results} reports these decisions and a stricter consent sensitivity analysis.
Complete-recovery rates use all scheduled runs as the denominator.

We record agent calls, tokens, simulated-user interactions and API expenditure from the checkpoint onward.
Costs are reported alongside outcomes, so unsuccessful early stops are not interpreted as efficient recovery.
Work to first correction is reported separately, conditional on reaching it.
Two conduct measures describe runs whether or not they recover. \emph{Acknowledged} records whether the agent tells the user about the error at any point after the checkpoint; it is judged from the agent's messages and requires a quoted message. \emph{Further changes} records, from the executed calls, whether a run that never corrects the error makes another change on the entity that the error affected; it does not judge whether that change is itself an error.
We also check whether a correction is subsequently undone, distinguishing temporary correction from correction preserved at termination.

\paragraph{Repairability over time.}
For selected source trajectories, we examine whether correction remains possible at successive states.
An exact closing point requires a demonstrated route immediately before an action and an impossibility certificate immediately after it.
This differs from undoing a correction, which may still be repairable.

\section{Results}
\label{sec:main-results}

\subsection{Evaluation Setup}
The main study evaluates 12 models: ten open-weight models and the hosted GPT-5.4-mini and GPT-6.1-sol.
Every model runs every $\tau^2$ and BFCL checkpoint under all four conditions, and the open-weight models also run AppWorld, with one run per checkpoint, model and condition: 5,760 continuations.
Open-weight models decode greedily.
Three supporting studies are reported in the appendix: a comparison that adds GPT-5.5 on the 37 $\tau^2$ and BFCL checkpoints with a demonstrated fix, repeated runs with three samples per cell on the same checkpoints, and competence controls on 18 BFCL checkpoints.
Appendix~\ref{app:experimental-setup} lists models and decoding settings, and Appendix~\ref{app:study-accounting} gives study populations, exclusions and run counts.

We report 95\% percentile confidence intervals from 2,000 bootstrap resamples of source tasks, retaining all checkpoints, models, conditions and repetitions within each sampled task to preserve matched comparisons.

\begin{table}[t]
\centering
\small
\begin{tabular}{@{}l@{\hspace{3pt}}r*{4}{@{\hspace{5pt}}r@{\hspace{3pt}}r}@{}}
\toprule
& & \multicolumn{2}{c}{U} & \multicolumn{2}{c}{N} & \multicolumn{2}{c}{L} & \multicolumn{2}{c}{PA} \\
\cmidrule(lr){3-4}\cmidrule(lr){5-6}\cmidrule(lr){7-8}\cmidrule(l){9-10}
Model & $n$ & C & \textbf{R} & C & \textbf{R} & C & \textbf{R} & C & \textbf{R} \\
\midrule
GPT-6.1-sol & 75 & .21 & .16 & .21 & .15 & .27 & .16 & .43 & .27 \\
Gemma-4-31B & 128 & .03 & .03 & .04 & .04 & .03 & .03 & .16 & .12 \\
Qwen3.5-35B-A3B & 128 & .02 & .01 & .05 & .04 & .02 & .02 & .12 & .09 \\
GPT-5.4-mini & 75 & .00 & .00 & .00 & .00 & .00 & .00 & .19 & .05 \\
gpt-oss-120b & 129 & .01 & .01 & .02 & .02 & .02 & .02 & .12 & .04 \\
Mistral-Small-3.2 & 129 & .01 & .00 & .01 & .01 & .01 & .01 & .11 & .04 \\
gpt-oss-20b & 129 & .00 & .00 & .01 & .01 & .01 & .01 & .05 & .03 \\
Llama-3.3-70B & 129 & .01 & .01 & .01 & .01 & .01 & .01 & .14 & .03 \\
Qwen3-235B & 128 & .01 & .01 & .00 & .00 & .00 & .00 & .05 & .02 \\
Qwen3-4B & 129 & .01 & .01 & .00 & .00 & .00 & .00 & .04 & .00 \\
Llama-3.1-8B & 129 & .00 & .00 & .01 & .01 & .00 & .00 & .02 & .00 \\
Qwen3-32B & 129 & .01 & .01 & .00 & .00 & .01 & .01 & .02 & .00 \\
\midrule
\emph{All models} & 1,437 & .02 & .02 & .02 & .02 & .02 & .02 & .11 & .05 \\
\bottomrule
\end{tabular}
\caption{Main evaluation on all 129 checkpoints, pooled over the environments each model was run in. C is the rate of runs that correct the error and R the rate of complete recoveries, both between 0 and 1, and $n$ is the number of runs per condition. U is unaided, N is notified, L is localized, and PA is plan-assisted. GPT-6.1-sol and GPT-5.4-mini ran on the 75 $\tau^2$-bench and BFCL checkpoints, and the open-weight models also on the 54 AppWorld checkpoints, with one run per checkpoint and condition; three cells are excluded (Appendix~\ref{app:study-accounting}). Rates include checkpoints with proven impossible or unknown fixability. Appendix Table~\ref{tab:appendix-full-grid} gives the counts in each environment, and Appendix Table~\ref{tab:scorecard} restricts the analysis to checkpoints with a demonstrated fix.}
\label{tab:main}
\end{table}

Table~\ref{tab:main} is the main evaluation: every model on every checkpoint under all four conditions, pooled over the environments the model was run in, with the rate of runs that correct the error and the rate that complete the recovery; Appendix Table~\ref{tab:appendix-full-grid} gives the counts in each environment.
Complete recovery is the primary outcome.
Figure~\ref{fig:map} and Appendix Table~\ref{tab:scorecard} are a diagnostic analysis on the 57 checkpoints with a demonstrated fix (Section~\ref{sec:diagnostic}).
AppWorld is run with the ten open-weight models only, and we do not combine the measures into one score.

\subsection{Recovery Is Hard, and Correcting Is Not Recovering}

Across the benchmark, complete recovery is rare.
Pooled over its environments (Table~\ref{tab:main}), the best model, GPT-6.1-sol, completes 16\% of recoveries unaided and 27\% with a plan.
Within a single environment (Appendix Table~\ref{tab:appendix-full-grid}), no model exceeds 21\% unaided, and with a repair plan the highest rate is GPT-6.1-sol's 34\% in $\tau^2$; the best rate is 24\% in BFCL, by Gemma-4-31B, and 11\% in AppWorld.
Summed over models, 17 of 455 plan-assisted runs are complete recoveries in $\tau^2$, 37 of 444 in BFCL and 18 of 538 in AppWorld.
Unaided, the counts are 8, 7 and 7, and all eight in $\tau^2$ are GPT-6.1-sol's.
These rates include the 15 checkpoints whose fix is proven impossible and the 57 for which none is known; Section~\ref{sec:diagnostic} looks at the errors that can be fixed.

Correction and complete recovery rank the models differently.
In $\tau^2$ with a plan, Llama-3.3-70B corrects 26\% of the errors and Mistral-Small-3.2 18\% without completing a recovery, and GPT-5.4-mini corrects 16\% and completes 3\%, while Qwen3.5-35B-A3B corrects 8\% and completes 5\%.
In BFCL, GPT-6.1-sol corrects more often than Gemma-4-31B (43\% against 30\%) but completes fewer recoveries (19\% against 24\%).
Size does not order the open-weight models either: in every environment Gemma-4-31B and Qwen3.5-35B-A3B complete more plan-assisted recoveries than Qwen3-235B, gpt-oss-120b and Llama-3.3-70B.

Task success does not track recovery.
Across the 38 $\tau^2$ checkpoints, Qwen3.5-35B-A3B passes the task evaluator in 14 to 18 runs in every condition and corrects no error without a plan.
In AppWorld, 40 of the 84 successful continuations leave the original error in place.

\subsection{Assistance Helps Different Models in Different Ways}

For most models a notice or the location leaves correction unchanged.
Summed over models, corrected runs go from 9 unaided to 11 with a notice and 14 with the location in $\tau^2$ (of 455), stay at 11 in BFCL (of 444), and go from 8 to 11 and 8 in AppWorld (of 538).
A plan raises them to 54, 72 and 25.
Against unaided continuation, a plan adds 10.0 percentage points of correction in $\tau^2$ ($[5.7,14.9]$), 13.7 in BFCL ($[6.8,21.2]$) and 3.3 in AppWorld ($[0.8,6.2]$), and 2.0 ($[0.9,3.2]$), 6.8 ($[2.4,11.7]$) and 2.1 ($[0.2,4.4]$) points of complete recovery.
The location points to the erroneous step in a history the agent can read, yet most models still do not correct it.
As in task-oriented dialogue \citep{dokme2026lost}, holding the information does not by itself update the agent's state of the task or govern its next action.
A plan supplies that update; without one, the agent must itself select the evidence in its history that bears on the error \citep{dokme-heck-2026-selective}.
Plan-assisted results measure recovery given repair guidance, not unprompted error detection.

The conditions do not act on every model in the same way.
Eleven models correct no $\tau^2$ error under the first three conditions, so for them only a plan matters.
For GPT-6.1-sol each step of assistance raises both correction (24\%, 29\%, 37\% and 42\%) and complete recovery (21\%, 24\%, 29\% and 34\%).

\paragraph{Telling the user is enough, and asking costs more.}
Told where the error is, GPT-6.1-sol corrects 14 $\tau^2$ errors and completes 11 recoveries; in nine of them it fixes the error first and tells the user afterwards, and in ten it never asks.
With a plan, which tells it to explain the error and ask before correcting, it completes 13 of its 16 corrections, all after asking.
Asking takes longer: its complete recoveries take a median of 15 agent calls with a plan, against 11 without one.
Three samples per checkpoint show the same ordering (Appendix Table~\ref{tab:main-outcomes}).

\begin{figure*}[t]
\centering
\resizebox{\textwidth}{!}{%
\begin{tikzpicture}[font=\footnotesize, x=1cm, y=1.25cm]
\node[anchor=east, inner sep=1pt] at (2.65,-0.135) {GPT-6.1-sol};
\node[anchor=east, inner sep=1pt] at (2.65,-0.405) {Gemma-4-31B};
\node[anchor=east, inner sep=1pt] at (2.65,-0.675) {Qwen3.5-35B-A3B};
\node[anchor=east, inner sep=1pt] at (2.65,-0.945) {GPT-5.4-mini};
\node[anchor=east, inner sep=1pt] at (2.65,-1.215) {gpt-oss-120b};
\node[anchor=east, inner sep=1pt] at (2.65,-1.485) {Mistral-Small-3.2};
\node[anchor=east, inner sep=1pt] at (2.65,-1.755) {gpt-oss-20b};
\node[anchor=east, inner sep=1pt] at (2.65,-2.025) {Llama-3.3-70B};
\node[anchor=east, inner sep=1pt] at (2.65,-2.295) {Qwen3-235B};
\node[anchor=east, inner sep=1pt] at (2.65,-2.565) {Qwen3-4B};
\node[anchor=east, inner sep=1pt] at (2.65,-2.835) {Llama-3.1-8B};
\node[anchor=east, inner sep=1pt] at (2.65,-3.105) {Qwen3-32B};
\node[anchor=south west, inner sep=1pt, font=\footnotesize\bfseries] at (2.75,0.03) {$\tau^2$-bench (18 checkpoints)};
\fill[visblue!35] (2.75,-0.250) rectangle ++(0.17,0.23);
\fill[visblue!90] (2.95,-0.250) rectangle ++(0.17,0.23);
\fill[visblue!90] (3.15,-0.250) rectangle ++(0.17,0.23);
\draw[black!25, line width=0.2pt] (3.35,-0.250) rectangle ++(0.17,0.23);
\draw[black!25, line width=0.2pt] (3.55,-0.250) rectangle ++(0.17,0.23);
\fill[visblue!90] (3.75,-0.250) rectangle ++(0.17,0.23);
\fill[visblue!90] (3.95,-0.250) rectangle ++(0.17,0.23);
\fill[visblue!35] (4.15,-0.250) rectangle ++(0.17,0.23);
\fill[visblue!90] (4.35,-0.250) rectangle ++(0.17,0.23);
\fill[visblue!90] (4.55,-0.250) rectangle ++(0.17,0.23);
\fill[visblue!90] (4.75,-0.250) rectangle ++(0.17,0.23);
\fill[visblue!35] (4.95,-0.250) rectangle ++(0.17,0.23);
\fill[visblue!90] (5.15,-0.250) rectangle ++(0.17,0.23);
\fill[visblue!90] (5.35,-0.250) rectangle ++(0.17,0.23);
\fill[visblue!90] (5.55,-0.250) rectangle ++(0.17,0.23);
\fill[visblue!90] (5.75,-0.250) rectangle ++(0.17,0.23);
\fill[visblue!90] (5.95,-0.250) rectangle ++(0.17,0.23);
\fill[visblue!90] (6.15,-0.250) rectangle ++(0.17,0.23);
\fill[visblue!35] (2.75,-0.520) rectangle ++(0.17,0.23);
\fill[visblue!90] (2.95,-0.520) rectangle ++(0.17,0.23);
\draw[black!25, line width=0.2pt] (3.15,-0.520) rectangle ++(0.17,0.23);
\draw[black!25, line width=0.2pt] (3.35,-0.520) rectangle ++(0.17,0.23);
\draw[black!25, line width=0.2pt] (3.55,-0.520) rectangle ++(0.17,0.23);
\draw[black!25, line width=0.2pt] (3.75,-0.520) rectangle ++(0.17,0.23);
\draw[black!25, line width=0.2pt] (3.95,-0.520) rectangle ++(0.17,0.23);
\draw[black!25, line width=0.2pt] (4.15,-0.520) rectangle ++(0.17,0.23);
\draw[black!25, line width=0.2pt] (4.35,-0.520) rectangle ++(0.17,0.23);
\draw[black!25, line width=0.2pt] (4.55,-0.520) rectangle ++(0.17,0.23);
\draw[black!25, line width=0.2pt] (4.75,-0.520) rectangle ++(0.17,0.23);
\draw[black!25, line width=0.2pt] (4.95,-0.520) rectangle ++(0.17,0.23);
\draw[black!25, line width=0.2pt] (5.15,-0.520) rectangle ++(0.17,0.23);
\draw[black!25, line width=0.2pt] (5.35,-0.520) rectangle ++(0.17,0.23);
\draw[black!25, line width=0.2pt] (5.55,-0.520) rectangle ++(0.17,0.23);
\draw[black!25, line width=0.2pt] (5.75,-0.520) rectangle ++(0.17,0.23);
\draw[black!25, line width=0.2pt] (5.95,-0.520) rectangle ++(0.17,0.23);
\draw[black!25, line width=0.2pt] (6.15,-0.520) rectangle ++(0.17,0.23);
\draw[black!25, line width=0.2pt] (2.75,-0.790) rectangle ++(0.17,0.23);
\draw[black!25, line width=0.2pt] (2.95,-0.790) rectangle ++(0.17,0.23);
\draw[black!25, line width=0.2pt] (3.15,-0.790) rectangle ++(0.17,0.23);
\draw[black!25, line width=0.2pt] (3.35,-0.790) rectangle ++(0.17,0.23);
\fill[visblue!35] (3.55,-0.790) rectangle ++(0.17,0.23);
\fill[visblue!90] (3.75,-0.790) rectangle ++(0.17,0.23);
\draw[black!25, line width=0.2pt] (3.95,-0.790) rectangle ++(0.17,0.23);
\fill[visblue!90] (4.15,-0.790) rectangle ++(0.17,0.23);
\draw[black!25, line width=0.2pt] (4.35,-0.790) rectangle ++(0.17,0.23);
\draw[black!25, line width=0.2pt] (4.55,-0.790) rectangle ++(0.17,0.23);
\draw[black!25, line width=0.2pt] (4.75,-0.790) rectangle ++(0.17,0.23);
\draw[black!25, line width=0.2pt] (4.95,-0.790) rectangle ++(0.17,0.23);
\draw[black!25, line width=0.2pt] (5.15,-0.790) rectangle ++(0.17,0.23);
\draw[black!25, line width=0.2pt] (5.35,-0.790) rectangle ++(0.17,0.23);
\draw[black!25, line width=0.2pt] (5.55,-0.790) rectangle ++(0.17,0.23);
\draw[black!25, line width=0.2pt] (5.75,-0.790) rectangle ++(0.17,0.23);
\draw[black!25, line width=0.2pt] (5.95,-0.790) rectangle ++(0.17,0.23);
\draw[black!25, line width=0.2pt] (6.15,-0.790) rectangle ++(0.17,0.23);
\fill[visblue!35] (2.75,-1.060) rectangle ++(0.17,0.23);
\draw[black!25, line width=0.2pt] (2.95,-1.060) rectangle ++(0.17,0.23);
\fill[visblue!35] (3.15,-1.060) rectangle ++(0.17,0.23);
\draw[black!25, line width=0.2pt] (3.35,-1.060) rectangle ++(0.17,0.23);
\draw[black!25, line width=0.2pt] (3.55,-1.060) rectangle ++(0.17,0.23);
\draw[black!25, line width=0.2pt] (3.75,-1.060) rectangle ++(0.17,0.23);
\fill[visblue!35] (3.95,-1.060) rectangle ++(0.17,0.23);
\draw[black!25, line width=0.2pt] (4.15,-1.060) rectangle ++(0.17,0.23);
\draw[black!25, line width=0.2pt] (4.35,-1.060) rectangle ++(0.17,0.23);
\draw[black!25, line width=0.2pt] (4.55,-1.060) rectangle ++(0.17,0.23);
\fill[visblue!35] (4.75,-1.060) rectangle ++(0.17,0.23);
\fill[visblue!35] (4.95,-1.060) rectangle ++(0.17,0.23);
\fill[visblue!90] (5.15,-1.060) rectangle ++(0.17,0.23);
\draw[black!25, line width=0.2pt] (5.35,-1.060) rectangle ++(0.17,0.23);
\draw[black!25, line width=0.2pt] (5.55,-1.060) rectangle ++(0.17,0.23);
\draw[black!25, line width=0.2pt] (5.75,-1.060) rectangle ++(0.17,0.23);
\draw[black!25, line width=0.2pt] (5.95,-1.060) rectangle ++(0.17,0.23);
\draw[black!25, line width=0.2pt] (6.15,-1.060) rectangle ++(0.17,0.23);
\fill[visblue!35] (2.75,-1.330) rectangle ++(0.17,0.23);
\fill[visblue!35] (2.95,-1.330) rectangle ++(0.17,0.23);
\draw[black!25, line width=0.2pt] (3.15,-1.330) rectangle ++(0.17,0.23);
\draw[black!25, line width=0.2pt] (3.35,-1.330) rectangle ++(0.17,0.23);
\fill[visblue!35] (3.55,-1.330) rectangle ++(0.17,0.23);
\draw[black!25, line width=0.2pt] (3.75,-1.330) rectangle ++(0.17,0.23);
\draw[black!25, line width=0.2pt] (3.95,-1.330) rectangle ++(0.17,0.23);
\draw[black!25, line width=0.2pt] (4.15,-1.330) rectangle ++(0.17,0.23);
\fill[visblue!35] (4.35,-1.330) rectangle ++(0.17,0.23);
\draw[black!25, line width=0.2pt] (4.55,-1.330) rectangle ++(0.17,0.23);
\draw[black!25, line width=0.2pt] (4.75,-1.330) rectangle ++(0.17,0.23);
\draw[black!25, line width=0.2pt] (4.95,-1.330) rectangle ++(0.17,0.23);
\draw[black!25, line width=0.2pt] (5.15,-1.330) rectangle ++(0.17,0.23);
\draw[black!25, line width=0.2pt] (5.35,-1.330) rectangle ++(0.17,0.23);
\draw[black!25, line width=0.2pt] (5.55,-1.330) rectangle ++(0.17,0.23);
\draw[black!25, line width=0.2pt] (5.75,-1.330) rectangle ++(0.17,0.23);
\draw[black!25, line width=0.2pt] (5.95,-1.330) rectangle ++(0.17,0.23);
\draw[black!25, line width=0.2pt] (6.15,-1.330) rectangle ++(0.17,0.23);
\fill[visblue!35] (2.75,-1.600) rectangle ++(0.17,0.23);
\fill[visblue!35] (2.95,-1.600) rectangle ++(0.17,0.23);
\fill[visblue!35] (3.15,-1.600) rectangle ++(0.17,0.23);
\draw[black!25, line width=0.2pt] (3.35,-1.600) rectangle ++(0.17,0.23);
\fill[visblue!35] (3.55,-1.600) rectangle ++(0.17,0.23);
\fill[visblue!35] (3.75,-1.600) rectangle ++(0.17,0.23);
\draw[black!25, line width=0.2pt] (3.95,-1.600) rectangle ++(0.17,0.23);
\draw[black!25, line width=0.2pt] (4.15,-1.600) rectangle ++(0.17,0.23);
\draw[black!25, line width=0.2pt] (4.35,-1.600) rectangle ++(0.17,0.23);
\draw[black!25, line width=0.2pt] (4.55,-1.600) rectangle ++(0.17,0.23);
\fill[visblue!35] (4.75,-1.600) rectangle ++(0.17,0.23);
\draw[black!25, line width=0.2pt] (4.95,-1.600) rectangle ++(0.17,0.23);
\draw[black!25, line width=0.2pt] (5.15,-1.600) rectangle ++(0.17,0.23);
\draw[black!25, line width=0.2pt] (5.35,-1.600) rectangle ++(0.17,0.23);
\fill[visblue!35] (5.55,-1.600) rectangle ++(0.17,0.23);
\draw[black!25, line width=0.2pt] (5.75,-1.600) rectangle ++(0.17,0.23);
\draw[black!25, line width=0.2pt] (5.95,-1.600) rectangle ++(0.17,0.23);
\draw[black!25, line width=0.2pt] (6.15,-1.600) rectangle ++(0.17,0.23);
\draw[black!25, line width=0.2pt] (2.75,-1.870) rectangle ++(0.17,0.23);
\draw[black!25, line width=0.2pt] (2.95,-1.870) rectangle ++(0.17,0.23);
\draw[black!25, line width=0.2pt] (3.15,-1.870) rectangle ++(0.17,0.23);
\draw[black!25, line width=0.2pt] (3.35,-1.870) rectangle ++(0.17,0.23);
\draw[black!25, line width=0.2pt] (3.55,-1.870) rectangle ++(0.17,0.23);
\draw[black!25, line width=0.2pt] (3.75,-1.870) rectangle ++(0.17,0.23);
\draw[black!25, line width=0.2pt] (3.95,-1.870) rectangle ++(0.17,0.23);
\draw[black!25, line width=0.2pt] (4.15,-1.870) rectangle ++(0.17,0.23);
\draw[black!25, line width=0.2pt] (4.35,-1.870) rectangle ++(0.17,0.23);
\draw[black!25, line width=0.2pt] (4.55,-1.870) rectangle ++(0.17,0.23);
\draw[black!25, line width=0.2pt] (4.75,-1.870) rectangle ++(0.17,0.23);
\draw[black!25, line width=0.2pt] (4.95,-1.870) rectangle ++(0.17,0.23);
\draw[black!25, line width=0.2pt] (5.15,-1.870) rectangle ++(0.17,0.23);
\draw[black!25, line width=0.2pt] (5.35,-1.870) rectangle ++(0.17,0.23);
\draw[black!25, line width=0.2pt] (5.55,-1.870) rectangle ++(0.17,0.23);
\draw[black!25, line width=0.2pt] (5.75,-1.870) rectangle ++(0.17,0.23);
\draw[black!25, line width=0.2pt] (5.95,-1.870) rectangle ++(0.17,0.23);
\draw[black!25, line width=0.2pt] (6.15,-1.870) rectangle ++(0.17,0.23);
\fill[visblue!35] (2.75,-2.140) rectangle ++(0.17,0.23);
\draw[black!25, line width=0.2pt] (2.95,-2.140) rectangle ++(0.17,0.23);
\fill[visblue!35] (3.15,-2.140) rectangle ++(0.17,0.23);
\fill[visblue!35] (3.35,-2.140) rectangle ++(0.17,0.23);
\draw[black!25, line width=0.2pt] (3.55,-2.140) rectangle ++(0.17,0.23);
\draw[black!25, line width=0.2pt] (3.75,-2.140) rectangle ++(0.17,0.23);
\fill[visblue!35] (3.95,-2.140) rectangle ++(0.17,0.23);
\fill[visblue!35] (4.15,-2.140) rectangle ++(0.17,0.23);
\fill[visblue!35] (4.35,-2.140) rectangle ++(0.17,0.23);
\fill[visblue!35] (4.55,-2.140) rectangle ++(0.17,0.23);
\draw[black!25, line width=0.2pt] (4.75,-2.140) rectangle ++(0.17,0.23);
\fill[visblue!35] (4.95,-2.140) rectangle ++(0.17,0.23);
\draw[black!25, line width=0.2pt] (5.15,-2.140) rectangle ++(0.17,0.23);
\fill[visblue!35] (5.35,-2.140) rectangle ++(0.17,0.23);
\draw[black!25, line width=0.2pt] (5.55,-2.140) rectangle ++(0.17,0.23);
\fill[visblue!35] (5.75,-2.140) rectangle ++(0.17,0.23);
\draw[black!25, line width=0.2pt] (5.95,-2.140) rectangle ++(0.17,0.23);
\draw[black!25, line width=0.2pt] (6.15,-2.140) rectangle ++(0.17,0.23);
\fill[visblue!35] (2.75,-2.410) rectangle ++(0.17,0.23);
\draw[black!25, line width=0.2pt] (2.95,-2.410) rectangle ++(0.17,0.23);
\draw[black!25, line width=0.2pt] (3.15,-2.410) rectangle ++(0.17,0.23);
\draw[black!25, line width=0.2pt] (3.35,-2.410) rectangle ++(0.17,0.23);
\draw[black!25, line width=0.2pt] (3.55,-2.410) rectangle ++(0.17,0.23);
\draw[black!25, line width=0.2pt] (3.75,-2.410) rectangle ++(0.17,0.23);
\draw[black!25, line width=0.2pt] (3.95,-2.410) rectangle ++(0.17,0.23);
\draw[black!25, line width=0.2pt] (4.15,-2.410) rectangle ++(0.17,0.23);
\draw[black!25, line width=0.2pt] (4.35,-2.410) rectangle ++(0.17,0.23);
\draw[black!25, line width=0.2pt] (4.55,-2.410) rectangle ++(0.17,0.23);
\draw[black!25, line width=0.2pt] (4.75,-2.410) rectangle ++(0.17,0.23);
\draw[black!25, line width=0.2pt] (4.95,-2.410) rectangle ++(0.17,0.23);
\draw[black!25, line width=0.2pt] (5.15,-2.410) rectangle ++(0.17,0.23);
\draw[black!25, line width=0.2pt] (5.35,-2.410) rectangle ++(0.17,0.23);
\draw[black!25, line width=0.2pt] (5.55,-2.410) rectangle ++(0.17,0.23);
\draw[black!25, line width=0.2pt] (5.75,-2.410) rectangle ++(0.17,0.23);
\draw[black!25, line width=0.2pt] (5.95,-2.410) rectangle ++(0.17,0.23);
\draw[black!25, line width=0.2pt] (6.15,-2.410) rectangle ++(0.17,0.23);
\draw[black!25, line width=0.2pt] (2.75,-2.680) rectangle ++(0.17,0.23);
\draw[black!25, line width=0.2pt] (2.95,-2.680) rectangle ++(0.17,0.23);
\draw[black!25, line width=0.2pt] (3.15,-2.680) rectangle ++(0.17,0.23);
\fill[visblue!35] (3.35,-2.680) rectangle ++(0.17,0.23);
\fill[visblue!35] (3.55,-2.680) rectangle ++(0.17,0.23);
\draw[black!25, line width=0.2pt] (3.75,-2.680) rectangle ++(0.17,0.23);
\draw[black!25, line width=0.2pt] (3.95,-2.680) rectangle ++(0.17,0.23);
\draw[black!25, line width=0.2pt] (4.15,-2.680) rectangle ++(0.17,0.23);
\draw[black!25, line width=0.2pt] (4.35,-2.680) rectangle ++(0.17,0.23);
\fill[visblue!35] (4.55,-2.680) rectangle ++(0.17,0.23);
\draw[black!25, line width=0.2pt] (4.75,-2.680) rectangle ++(0.17,0.23);
\draw[black!25, line width=0.2pt] (4.95,-2.680) rectangle ++(0.17,0.23);
\draw[black!25, line width=0.2pt] (5.15,-2.680) rectangle ++(0.17,0.23);
\draw[black!25, line width=0.2pt] (5.35,-2.680) rectangle ++(0.17,0.23);
\draw[black!25, line width=0.2pt] (5.55,-2.680) rectangle ++(0.17,0.23);
\draw[black!25, line width=0.2pt] (5.75,-2.680) rectangle ++(0.17,0.23);
\draw[black!25, line width=0.2pt] (5.95,-2.680) rectangle ++(0.17,0.23);
\draw[black!25, line width=0.2pt] (6.15,-2.680) rectangle ++(0.17,0.23);
\draw[black!25, line width=0.2pt] (2.75,-2.950) rectangle ++(0.17,0.23);
\draw[black!25, line width=0.2pt] (2.95,-2.950) rectangle ++(0.17,0.23);
\draw[black!25, line width=0.2pt] (3.15,-2.950) rectangle ++(0.17,0.23);
\fill[visblue!35] (3.35,-2.950) rectangle ++(0.17,0.23);
\draw[black!25, line width=0.2pt] (3.55,-2.950) rectangle ++(0.17,0.23);
\draw[black!25, line width=0.2pt] (3.75,-2.950) rectangle ++(0.17,0.23);
\draw[black!25, line width=0.2pt] (3.95,-2.950) rectangle ++(0.17,0.23);
\draw[black!25, line width=0.2pt] (4.15,-2.950) rectangle ++(0.17,0.23);
\draw[black!25, line width=0.2pt] (4.35,-2.950) rectangle ++(0.17,0.23);
\draw[black!25, line width=0.2pt] (4.55,-2.950) rectangle ++(0.17,0.23);
\draw[black!25, line width=0.2pt] (4.75,-2.950) rectangle ++(0.17,0.23);
\draw[black!25, line width=0.2pt] (4.95,-2.950) rectangle ++(0.17,0.23);
\draw[black!25, line width=0.2pt] (5.15,-2.950) rectangle ++(0.17,0.23);
\draw[black!25, line width=0.2pt] (5.35,-2.950) rectangle ++(0.17,0.23);
\draw[black!25, line width=0.2pt] (5.55,-2.950) rectangle ++(0.17,0.23);
\draw[black!25, line width=0.2pt] (5.75,-2.950) rectangle ++(0.17,0.23);
\draw[black!25, line width=0.2pt] (5.95,-2.950) rectangle ++(0.17,0.23);
\draw[black!25, line width=0.2pt] (6.15,-2.950) rectangle ++(0.17,0.23);
\draw[black!25, line width=0.2pt] (2.75,-3.220) rectangle ++(0.17,0.23);
\draw[black!25, line width=0.2pt] (2.95,-3.220) rectangle ++(0.17,0.23);
\draw[black!25, line width=0.2pt] (3.15,-3.220) rectangle ++(0.17,0.23);
\fill[visblue!35] (3.35,-3.220) rectangle ++(0.17,0.23);
\draw[black!25, line width=0.2pt] (3.55,-3.220) rectangle ++(0.17,0.23);
\draw[black!25, line width=0.2pt] (3.75,-3.220) rectangle ++(0.17,0.23);
\draw[black!25, line width=0.2pt] (3.95,-3.220) rectangle ++(0.17,0.23);
\draw[black!25, line width=0.2pt] (4.15,-3.220) rectangle ++(0.17,0.23);
\draw[black!25, line width=0.2pt] (4.35,-3.220) rectangle ++(0.17,0.23);
\draw[black!25, line width=0.2pt] (4.55,-3.220) rectangle ++(0.17,0.23);
\draw[black!25, line width=0.2pt] (4.75,-3.220) rectangle ++(0.17,0.23);
\draw[black!25, line width=0.2pt] (4.95,-3.220) rectangle ++(0.17,0.23);
\draw[black!25, line width=0.2pt] (5.15,-3.220) rectangle ++(0.17,0.23);
\draw[black!25, line width=0.2pt] (5.35,-3.220) rectangle ++(0.17,0.23);
\draw[black!25, line width=0.2pt] (5.55,-3.220) rectangle ++(0.17,0.23);
\draw[black!25, line width=0.2pt] (5.75,-3.220) rectangle ++(0.17,0.23);
\draw[black!25, line width=0.2pt] (5.95,-3.220) rectangle ++(0.17,0.23);
\draw[black!25, line width=0.2pt] (6.15,-3.220) rectangle ++(0.17,0.23);
\node[anchor=south west, inner sep=1pt, font=\footnotesize\bfseries] at (6.70,0.03) {BFCL (19 checkpoints)};
\fill[visblue!35] (6.70,-0.250) rectangle ++(0.17,0.23);
\fill[visblue!35] (6.90,-0.250) rectangle ++(0.17,0.23);
\fill[visblue!35] (7.10,-0.250) rectangle ++(0.17,0.23);
\fill[visblue!90] (7.30,-0.250) rectangle ++(0.17,0.23);
\fill[visblue!35] (7.50,-0.250) rectangle ++(0.17,0.23);
\fill[visblue!35] (7.70,-0.250) rectangle ++(0.17,0.23);
\fill[visblue!90] (7.90,-0.250) rectangle ++(0.17,0.23);
\fill[visblue!90] (8.10,-0.250) rectangle ++(0.17,0.23);
\fill[visblue!35] (8.30,-0.250) rectangle ++(0.17,0.23);
\fill[visblue!90] (8.50,-0.250) rectangle ++(0.17,0.23);
\fill[visblue!90] (8.70,-0.250) rectangle ++(0.17,0.23);
\fill[visblue!90] (8.90,-0.250) rectangle ++(0.17,0.23);
\fill[visblue!90] (9.10,-0.250) rectangle ++(0.17,0.23);
\draw[black!25, line width=0.2pt] (9.30,-0.250) rectangle ++(0.17,0.23);
\fill[visblue!35] (9.50,-0.250) rectangle ++(0.17,0.23);
\draw[black!25, line width=0.2pt] (9.70,-0.250) rectangle ++(0.17,0.23);
\draw[black!25, line width=0.2pt] (9.90,-0.250) rectangle ++(0.17,0.23);
\draw[black!25, line width=0.2pt] (10.10,-0.250) rectangle ++(0.17,0.23);
\draw[black!25, line width=0.2pt] (10.30,-0.250) rectangle ++(0.17,0.23);
\fill[visblue!90] (6.70,-0.520) rectangle ++(0.17,0.23);
\fill[visblue!90] (6.90,-0.520) rectangle ++(0.17,0.23);
\fill[visblue!90] (7.10,-0.520) rectangle ++(0.17,0.23);
\fill[visblue!90] (7.30,-0.520) rectangle ++(0.17,0.23);
\fill[visblue!90] (7.50,-0.520) rectangle ++(0.17,0.23);
\fill[visblue!35] (7.70,-0.520) rectangle ++(0.17,0.23);
\fill[visblue!35] (7.90,-0.520) rectangle ++(0.17,0.23);
\fill[visblue!90] (8.10,-0.520) rectangle ++(0.17,0.23);
\fill[visblue!90] (8.30,-0.520) rectangle ++(0.17,0.23);
\draw[black!25, line width=0.2pt] (8.50,-0.520) rectangle ++(0.17,0.23);
\fill[visblue!90] (8.70,-0.520) rectangle ++(0.17,0.23);
\fill[visblue!90] (8.90,-0.520) rectangle ++(0.17,0.23);
\draw[black!25, line width=0.2pt] (9.10,-0.520) rectangle ++(0.17,0.23);
\draw[black!25, line width=0.2pt] (9.30,-0.520) rectangle ++(0.17,0.23);
\draw[black!25, line width=0.2pt] (9.50,-0.520) rectangle ++(0.17,0.23);
\draw[black!25, line width=0.2pt] (9.70,-0.520) rectangle ++(0.17,0.23);
\draw[black!25, line width=0.2pt] (9.90,-0.520) rectangle ++(0.17,0.23);
\draw[black!25, line width=0.2pt] (10.10,-0.520) rectangle ++(0.17,0.23);
\draw[black!25, line width=0.2pt] (10.30,-0.520) rectangle ++(0.17,0.23);
\fill[visblue!90] (6.70,-0.790) rectangle ++(0.17,0.23);
\fill[visblue!90] (6.90,-0.790) rectangle ++(0.17,0.23);
\fill[visblue!90] (7.10,-0.790) rectangle ++(0.17,0.23);
\fill[visblue!90] (7.30,-0.790) rectangle ++(0.17,0.23);
\fill[visblue!90] (7.50,-0.790) rectangle ++(0.17,0.23);
\draw[black!25, line width=0.2pt] (7.70,-0.790) rectangle ++(0.17,0.23);
\draw[black!25, line width=0.2pt] (7.90,-0.790) rectangle ++(0.17,0.23);
\fill[visblue!35] (8.10,-0.790) rectangle ++(0.17,0.23);
\draw[black!25, line width=0.2pt] (8.30,-0.790) rectangle ++(0.17,0.23);
\draw[black!25, line width=0.2pt] (8.50,-0.790) rectangle ++(0.17,0.23);
\draw[black!25, line width=0.2pt] (8.70,-0.790) rectangle ++(0.17,0.23);
\fill[visblue!35] (8.90,-0.790) rectangle ++(0.17,0.23);
\draw[black!25, line width=0.2pt] (9.10,-0.790) rectangle ++(0.17,0.23);
\draw[black!25, line width=0.2pt] (9.30,-0.790) rectangle ++(0.17,0.23);
\draw[black!25, line width=0.2pt] (9.50,-0.790) rectangle ++(0.17,0.23);
\draw[black!25, line width=0.2pt] (9.70,-0.790) rectangle ++(0.17,0.23);
\draw[black!25, line width=0.2pt] (9.90,-0.790) rectangle ++(0.17,0.23);
\draw[black!25, line width=0.2pt] (10.10,-0.790) rectangle ++(0.17,0.23);
\draw[black!25, line width=0.2pt] (10.30,-0.790) rectangle ++(0.17,0.23);
\fill[visblue!35] (6.70,-1.060) rectangle ++(0.17,0.23);
\fill[visblue!90] (6.90,-1.060) rectangle ++(0.17,0.23);
\fill[visblue!90] (7.10,-1.060) rectangle ++(0.17,0.23);
\fill[visblue!35] (7.30,-1.060) rectangle ++(0.17,0.23);
\fill[visblue!35] (7.50,-1.060) rectangle ++(0.17,0.23);
\fill[visblue!35] (7.70,-1.060) rectangle ++(0.17,0.23);
\draw[black!25, line width=0.2pt] (7.90,-1.060) rectangle ++(0.17,0.23);
\draw[black!25, line width=0.2pt] (8.10,-1.060) rectangle ++(0.17,0.23);
\draw[black!25, line width=0.2pt] (8.30,-1.060) rectangle ++(0.17,0.23);
\draw[black!25, line width=0.2pt] (8.50,-1.060) rectangle ++(0.17,0.23);
\draw[black!25, line width=0.2pt] (8.70,-1.060) rectangle ++(0.17,0.23);
\draw[black!25, line width=0.2pt] (8.90,-1.060) rectangle ++(0.17,0.23);
\fill[visblue!90] (9.10,-1.060) rectangle ++(0.17,0.23);
\fill[visblue!35] (9.30,-1.060) rectangle ++(0.17,0.23);
\draw[black!25, line width=0.2pt] (9.50,-1.060) rectangle ++(0.17,0.23);
\draw[black!25, line width=0.2pt] (9.70,-1.060) rectangle ++(0.17,0.23);
\draw[black!25, line width=0.2pt] (9.90,-1.060) rectangle ++(0.17,0.23);
\draw[black!25, line width=0.2pt] (10.10,-1.060) rectangle ++(0.17,0.23);
\draw[black!25, line width=0.2pt] (10.30,-1.060) rectangle ++(0.17,0.23);
\fill[visblue!90] (6.70,-1.330) rectangle ++(0.17,0.23);
\fill[visblue!90] (6.90,-1.330) rectangle ++(0.17,0.23);
\fill[visblue!35] (7.10,-1.330) rectangle ++(0.17,0.23);
\fill[visblue!35] (7.30,-1.330) rectangle ++(0.17,0.23);
\draw[black!25, line width=0.2pt] (7.50,-1.330) rectangle ++(0.17,0.23);
\fill[visblue!35] (7.70,-1.330) rectangle ++(0.17,0.23);
\fill[visblue!90] (7.90,-1.330) rectangle ++(0.17,0.23);
\draw[black!25, line width=0.2pt] (8.10,-1.330) rectangle ++(0.17,0.23);
\fill[visblue!35] (8.30,-1.330) rectangle ++(0.17,0.23);
\fill[visblue!90] (8.50,-1.330) rectangle ++(0.17,0.23);
\draw[black!25, line width=0.2pt] (8.70,-1.330) rectangle ++(0.17,0.23);
\draw[black!25, line width=0.2pt] (8.90,-1.330) rectangle ++(0.17,0.23);
\draw[black!25, line width=0.2pt] (9.10,-1.330) rectangle ++(0.17,0.23);
\fill[visblue!35] (9.30,-1.330) rectangle ++(0.17,0.23);
\draw[black!25, line width=0.2pt] (9.50,-1.330) rectangle ++(0.17,0.23);
\draw[black!25, line width=0.2pt] (9.70,-1.330) rectangle ++(0.17,0.23);
\draw[black!25, line width=0.2pt] (9.90,-1.330) rectangle ++(0.17,0.23);
\draw[black!25, line width=0.2pt] (10.10,-1.330) rectangle ++(0.17,0.23);
\draw[black!25, line width=0.2pt] (10.30,-1.330) rectangle ++(0.17,0.23);
\draw[black!25, line width=0.2pt] (6.70,-1.600) rectangle ++(0.17,0.23);
\fill[visblue!90] (6.90,-1.600) rectangle ++(0.17,0.23);
\fill[visblue!90] (7.10,-1.600) rectangle ++(0.17,0.23);
\draw[black!25, line width=0.2pt] (7.30,-1.600) rectangle ++(0.17,0.23);
\draw[black!25, line width=0.2pt] (7.50,-1.600) rectangle ++(0.17,0.23);
\draw[black!25, line width=0.2pt] (7.70,-1.600) rectangle ++(0.17,0.23);
\fill[visblue!90] (7.90,-1.600) rectangle ++(0.17,0.23);
\draw[black!25, line width=0.2pt] (8.10,-1.600) rectangle ++(0.17,0.23);
\draw[black!25, line width=0.2pt] (8.30,-1.600) rectangle ++(0.17,0.23);
\fill[visblue!90] (8.50,-1.600) rectangle ++(0.17,0.23);
\draw[black!25, line width=0.2pt] (8.70,-1.600) rectangle ++(0.17,0.23);
\draw[black!25, line width=0.2pt] (8.90,-1.600) rectangle ++(0.17,0.23);
\draw[black!25, line width=0.2pt] (9.10,-1.600) rectangle ++(0.17,0.23);
\draw[black!25, line width=0.2pt] (9.30,-1.600) rectangle ++(0.17,0.23);
\draw[black!25, line width=0.2pt] (9.50,-1.600) rectangle ++(0.17,0.23);
\draw[black!25, line width=0.2pt] (9.70,-1.600) rectangle ++(0.17,0.23);
\draw[black!25, line width=0.2pt] (9.90,-1.600) rectangle ++(0.17,0.23);
\draw[black!25, line width=0.2pt] (10.10,-1.600) rectangle ++(0.17,0.23);
\draw[black!25, line width=0.2pt] (10.30,-1.600) rectangle ++(0.17,0.23);
\fill[visblue!90] (6.70,-1.870) rectangle ++(0.17,0.23);
\fill[visblue!90] (6.90,-1.870) rectangle ++(0.17,0.23);
\fill[visblue!90] (7.10,-1.870) rectangle ++(0.17,0.23);
\fill[visblue!35] (7.30,-1.870) rectangle ++(0.17,0.23);
\fill[visblue!35] (7.50,-1.870) rectangle ++(0.17,0.23);
\draw[black!25, line width=0.2pt] (7.70,-1.870) rectangle ++(0.17,0.23);
\draw[black!25, line width=0.2pt] (7.90,-1.870) rectangle ++(0.17,0.23);
\draw[black!25, line width=0.2pt] (8.10,-1.870) rectangle ++(0.17,0.23);
\draw[black!25, line width=0.2pt] (8.30,-1.870) rectangle ++(0.17,0.23);
\draw[black!25, line width=0.2pt] (8.50,-1.870) rectangle ++(0.17,0.23);
\draw[black!25, line width=0.2pt] (8.70,-1.870) rectangle ++(0.17,0.23);
\draw[black!25, line width=0.2pt] (8.90,-1.870) rectangle ++(0.17,0.23);
\draw[black!25, line width=0.2pt] (9.10,-1.870) rectangle ++(0.17,0.23);
\draw[black!25, line width=0.2pt] (9.30,-1.870) rectangle ++(0.17,0.23);
\draw[black!25, line width=0.2pt] (9.50,-1.870) rectangle ++(0.17,0.23);
\fill[visblue!35] (9.70,-1.870) rectangle ++(0.17,0.23);
\draw[black!25, line width=0.2pt] (9.90,-1.870) rectangle ++(0.17,0.23);
\draw[black!25, line width=0.2pt] (10.10,-1.870) rectangle ++(0.17,0.23);
\draw[black!25, line width=0.2pt] (10.30,-1.870) rectangle ++(0.17,0.23);
\fill[visblue!90] (6.70,-2.140) rectangle ++(0.17,0.23);
\draw[black!25, line width=0.2pt] (6.90,-2.140) rectangle ++(0.17,0.23);
\draw[black!25, line width=0.2pt] (7.10,-2.140) rectangle ++(0.17,0.23);
\draw[black!25, line width=0.2pt] (7.30,-2.140) rectangle ++(0.17,0.23);
\draw[black!25, line width=0.2pt] (7.50,-2.140) rectangle ++(0.17,0.23);
\draw[black!25, line width=0.2pt] (7.70,-2.140) rectangle ++(0.17,0.23);
\draw[black!25, line width=0.2pt] (7.90,-2.140) rectangle ++(0.17,0.23);
\fill[visblue!35] (8.10,-2.140) rectangle ++(0.17,0.23);
\fill[visblue!35] (8.30,-2.140) rectangle ++(0.17,0.23);
\draw[black!25, line width=0.2pt] (8.50,-2.140) rectangle ++(0.17,0.23);
\fill[visblue!35] (8.70,-2.140) rectangle ++(0.17,0.23);
\draw[black!25, line width=0.2pt] (8.90,-2.140) rectangle ++(0.17,0.23);
\draw[black!25, line width=0.2pt] (9.10,-2.140) rectangle ++(0.17,0.23);
\draw[black!25, line width=0.2pt] (9.30,-2.140) rectangle ++(0.17,0.23);
\draw[black!25, line width=0.2pt] (9.50,-2.140) rectangle ++(0.17,0.23);
\draw[black!25, line width=0.2pt] (9.70,-2.140) rectangle ++(0.17,0.23);
\draw[black!25, line width=0.2pt] (9.90,-2.140) rectangle ++(0.17,0.23);
\draw[black!25, line width=0.2pt] (10.10,-2.140) rectangle ++(0.17,0.23);
\draw[black!25, line width=0.2pt] (10.30,-2.140) rectangle ++(0.17,0.23);
\draw[black!25, line width=0.2pt] (6.70,-2.410) rectangle ++(0.17,0.23);
\fill[visblue!90] (6.90,-2.410) rectangle ++(0.17,0.23);
\draw[black!25, line width=0.2pt] (7.10,-2.410) rectangle ++(0.17,0.23);
\fill[visblue!35] (7.30,-2.410) rectangle ++(0.17,0.23);
\draw[black!25, line width=0.2pt] (7.50,-2.410) rectangle ++(0.17,0.23);
\fill[visblue!35] (7.70,-2.410) rectangle ++(0.17,0.23);
\draw[black!25, line width=0.2pt] (7.90,-2.410) rectangle ++(0.17,0.23);
\draw[black!25, line width=0.2pt] (8.10,-2.410) rectangle ++(0.17,0.23);
\draw[black!25, line width=0.2pt] (8.30,-2.410) rectangle ++(0.17,0.23);
\draw[black!25, line width=0.2pt] (8.50,-2.410) rectangle ++(0.17,0.23);
\draw[black!25, line width=0.2pt] (8.70,-2.410) rectangle ++(0.17,0.23);
\draw[black!25, line width=0.2pt] (8.90,-2.410) rectangle ++(0.17,0.23);
\draw[black!25, line width=0.2pt] (9.10,-2.410) rectangle ++(0.17,0.23);
\draw[black!25, line width=0.2pt] (9.30,-2.410) rectangle ++(0.17,0.23);
\draw[black!25, line width=0.2pt] (9.50,-2.410) rectangle ++(0.17,0.23);
\draw[black!25, line width=0.2pt] (9.70,-2.410) rectangle ++(0.17,0.23);
\draw[black!25, line width=0.2pt] (9.90,-2.410) rectangle ++(0.17,0.23);
\draw[black!25, line width=0.2pt] (10.10,-2.410) rectangle ++(0.17,0.23);
\draw[black!25, line width=0.2pt] (10.30,-2.410) rectangle ++(0.17,0.23);
\fill[visblue!35] (6.70,-2.680) rectangle ++(0.17,0.23);
\draw[black!25, line width=0.2pt] (6.90,-2.680) rectangle ++(0.17,0.23);
\draw[black!25, line width=0.2pt] (7.10,-2.680) rectangle ++(0.17,0.23);
\draw[black!25, line width=0.2pt] (7.30,-2.680) rectangle ++(0.17,0.23);
\draw[black!25, line width=0.2pt] (7.50,-2.680) rectangle ++(0.17,0.23);
\fill[visblue!35] (7.70,-2.680) rectangle ++(0.17,0.23);
\draw[black!25, line width=0.2pt] (7.90,-2.680) rectangle ++(0.17,0.23);
\draw[black!25, line width=0.2pt] (8.10,-2.680) rectangle ++(0.17,0.23);
\draw[black!25, line width=0.2pt] (8.30,-2.680) rectangle ++(0.17,0.23);
\draw[black!25, line width=0.2pt] (8.50,-2.680) rectangle ++(0.17,0.23);
\draw[black!25, line width=0.2pt] (8.70,-2.680) rectangle ++(0.17,0.23);
\draw[black!25, line width=0.2pt] (8.90,-2.680) rectangle ++(0.17,0.23);
\draw[black!25, line width=0.2pt] (9.10,-2.680) rectangle ++(0.17,0.23);
\draw[black!25, line width=0.2pt] (9.30,-2.680) rectangle ++(0.17,0.23);
\draw[black!25, line width=0.2pt] (9.50,-2.680) rectangle ++(0.17,0.23);
\draw[black!25, line width=0.2pt] (9.70,-2.680) rectangle ++(0.17,0.23);
\draw[black!25, line width=0.2pt] (9.90,-2.680) rectangle ++(0.17,0.23);
\draw[black!25, line width=0.2pt] (10.10,-2.680) rectangle ++(0.17,0.23);
\draw[black!25, line width=0.2pt] (10.30,-2.680) rectangle ++(0.17,0.23);
\fill[visblue!35] (6.70,-2.950) rectangle ++(0.17,0.23);
\draw[black!25, line width=0.2pt] (6.90,-2.950) rectangle ++(0.17,0.23);
\draw[black!25, line width=0.2pt] (7.10,-2.950) rectangle ++(0.17,0.23);
\draw[black!25, line width=0.2pt] (7.30,-2.950) rectangle ++(0.17,0.23);
\draw[black!25, line width=0.2pt] (7.50,-2.950) rectangle ++(0.17,0.23);
\draw[black!25, line width=0.2pt] (7.70,-2.950) rectangle ++(0.17,0.23);
\draw[black!25, line width=0.2pt] (7.90,-2.950) rectangle ++(0.17,0.23);
\draw[black!25, line width=0.2pt] (8.10,-2.950) rectangle ++(0.17,0.23);
\draw[black!25, line width=0.2pt] (8.30,-2.950) rectangle ++(0.17,0.23);
\draw[black!25, line width=0.2pt] (8.50,-2.950) rectangle ++(0.17,0.23);
\draw[black!25, line width=0.2pt] (8.70,-2.950) rectangle ++(0.17,0.23);
\draw[black!25, line width=0.2pt] (8.90,-2.950) rectangle ++(0.17,0.23);
\draw[black!25, line width=0.2pt] (9.10,-2.950) rectangle ++(0.17,0.23);
\draw[black!25, line width=0.2pt] (9.30,-2.950) rectangle ++(0.17,0.23);
\draw[black!25, line width=0.2pt] (9.50,-2.950) rectangle ++(0.17,0.23);
\draw[black!25, line width=0.2pt] (9.70,-2.950) rectangle ++(0.17,0.23);
\draw[black!25, line width=0.2pt] (9.90,-2.950) rectangle ++(0.17,0.23);
\draw[black!25, line width=0.2pt] (10.10,-2.950) rectangle ++(0.17,0.23);
\draw[black!25, line width=0.2pt] (10.30,-2.950) rectangle ++(0.17,0.23);
\draw[black!25, line width=0.2pt] (6.70,-3.220) rectangle ++(0.17,0.23);
\draw[black!25, line width=0.2pt] (6.90,-3.220) rectangle ++(0.17,0.23);
\draw[black!25, line width=0.2pt] (7.10,-3.220) rectangle ++(0.17,0.23);
\draw[black!25, line width=0.2pt] (7.30,-3.220) rectangle ++(0.17,0.23);
\fill[visblue!35] (7.50,-3.220) rectangle ++(0.17,0.23);
\draw[black!25, line width=0.2pt] (7.70,-3.220) rectangle ++(0.17,0.23);
\draw[black!25, line width=0.2pt] (7.90,-3.220) rectangle ++(0.17,0.23);
\draw[black!25, line width=0.2pt] (8.10,-3.220) rectangle ++(0.17,0.23);
\draw[black!25, line width=0.2pt] (8.30,-3.220) rectangle ++(0.17,0.23);
\draw[black!25, line width=0.2pt] (8.50,-3.220) rectangle ++(0.17,0.23);
\draw[black!25, line width=0.2pt] (8.70,-3.220) rectangle ++(0.17,0.23);
\draw[black!25, line width=0.2pt] (8.90,-3.220) rectangle ++(0.17,0.23);
\draw[black!25, line width=0.2pt] (9.10,-3.220) rectangle ++(0.17,0.23);
\draw[black!25, line width=0.2pt] (9.30,-3.220) rectangle ++(0.17,0.23);
\draw[black!25, line width=0.2pt] (9.50,-3.220) rectangle ++(0.17,0.23);
\draw[black!25, line width=0.2pt] (9.70,-3.220) rectangle ++(0.17,0.23);
\draw[black!25, line width=0.2pt] (9.90,-3.220) rectangle ++(0.17,0.23);
\draw[black!25, line width=0.2pt] (10.10,-3.220) rectangle ++(0.17,0.23);
\draw[black!25, line width=0.2pt] (10.30,-3.220) rectangle ++(0.17,0.23);
\node[anchor=south west, inner sep=1pt, font=\footnotesize\bfseries] at (10.85,0.03) {AppWorld (20 checkpoints)};
\fill[black!14] (10.85,-0.250) rectangle ++(0.17,0.23);
\fill[black!14] (11.05,-0.250) rectangle ++(0.17,0.23);
\fill[black!14] (11.25,-0.250) rectangle ++(0.17,0.23);
\fill[black!14] (11.45,-0.250) rectangle ++(0.17,0.23);
\fill[black!14] (11.65,-0.250) rectangle ++(0.17,0.23);
\fill[black!14] (11.85,-0.250) rectangle ++(0.17,0.23);
\fill[black!14] (12.05,-0.250) rectangle ++(0.17,0.23);
\fill[black!14] (12.25,-0.250) rectangle ++(0.17,0.23);
\fill[black!14] (12.45,-0.250) rectangle ++(0.17,0.23);
\fill[black!14] (12.65,-0.250) rectangle ++(0.17,0.23);
\fill[black!14] (12.85,-0.250) rectangle ++(0.17,0.23);
\fill[black!14] (13.05,-0.250) rectangle ++(0.17,0.23);
\fill[black!14] (13.25,-0.250) rectangle ++(0.17,0.23);
\fill[black!14] (13.45,-0.250) rectangle ++(0.17,0.23);
\fill[black!14] (13.65,-0.250) rectangle ++(0.17,0.23);
\fill[black!14] (13.85,-0.250) rectangle ++(0.17,0.23);
\fill[black!14] (14.05,-0.250) rectangle ++(0.17,0.23);
\fill[black!14] (14.25,-0.250) rectangle ++(0.17,0.23);
\fill[black!14] (14.45,-0.250) rectangle ++(0.17,0.23);
\fill[black!14] (14.65,-0.250) rectangle ++(0.17,0.23);
\fill[visblue!90] (10.85,-0.520) rectangle ++(0.17,0.23);
\fill[visblue!90] (11.05,-0.520) rectangle ++(0.17,0.23);
\fill[visblue!90] (11.25,-0.520) rectangle ++(0.17,0.23);
\fill[visblue!90] (11.45,-0.520) rectangle ++(0.17,0.23);
\fill[visblue!90] (11.65,-0.520) rectangle ++(0.17,0.23);
\draw[black!25, line width=0.2pt] (11.85,-0.520) rectangle ++(0.17,0.23);
\fill[visblue!90] (12.05,-0.520) rectangle ++(0.17,0.23);
\draw[black!25, line width=0.2pt] (12.25,-0.520) rectangle ++(0.17,0.23);
\draw[black!25, line width=0.2pt] (12.45,-0.520) rectangle ++(0.17,0.23);
\draw[black!25, line width=0.2pt] (12.65,-0.520) rectangle ++(0.17,0.23);
\fill[visblue!35] (12.85,-0.520) rectangle ++(0.17,0.23);
\draw[black!25, line width=0.2pt] (13.05,-0.520) rectangle ++(0.17,0.23);
\draw[black!25, line width=0.2pt] (13.25,-0.520) rectangle ++(0.17,0.23);
\draw[black!25, line width=0.2pt] (13.45,-0.520) rectangle ++(0.17,0.23);
\draw[black!25, line width=0.2pt] (13.65,-0.520) rectangle ++(0.17,0.23);
\draw[black!25, line width=0.2pt] (13.85,-0.520) rectangle ++(0.17,0.23);
\draw[black!25, line width=0.2pt] (14.05,-0.520) rectangle ++(0.17,0.23);
\draw[black!25, line width=0.2pt] (14.25,-0.520) rectangle ++(0.17,0.23);
\draw[black!25, line width=0.2pt] (14.45,-0.520) rectangle ++(0.17,0.23);
\draw[black!25, line width=0.2pt] (14.65,-0.520) rectangle ++(0.17,0.23);
\fill[visblue!90] (10.85,-0.790) rectangle ++(0.17,0.23);
\fill[visblue!90] (11.05,-0.790) rectangle ++(0.17,0.23);
\fill[visblue!90] (11.25,-0.790) rectangle ++(0.17,0.23);
\fill[visblue!90] (11.45,-0.790) rectangle ++(0.17,0.23);
\fill[visblue!90] (11.65,-0.790) rectangle ++(0.17,0.23);
\draw[black!25, line width=0.2pt] (11.85,-0.790) rectangle ++(0.17,0.23);
\draw[black!25, line width=0.2pt] (12.05,-0.790) rectangle ++(0.17,0.23);
\draw[black!25, line width=0.2pt] (12.25,-0.790) rectangle ++(0.17,0.23);
\draw[black!25, line width=0.2pt] (12.45,-0.790) rectangle ++(0.17,0.23);
\draw[black!25, line width=0.2pt] (12.65,-0.790) rectangle ++(0.17,0.23);
\draw[black!25, line width=0.2pt] (12.85,-0.790) rectangle ++(0.17,0.23);
\draw[black!25, line width=0.2pt] (13.05,-0.790) rectangle ++(0.17,0.23);
\draw[black!25, line width=0.2pt] (13.25,-0.790) rectangle ++(0.17,0.23);
\draw[black!25, line width=0.2pt] (13.45,-0.790) rectangle ++(0.17,0.23);
\draw[black!25, line width=0.2pt] (13.65,-0.790) rectangle ++(0.17,0.23);
\draw[black!25, line width=0.2pt] (13.85,-0.790) rectangle ++(0.17,0.23);
\draw[black!25, line width=0.2pt] (14.05,-0.790) rectangle ++(0.17,0.23);
\draw[black!25, line width=0.2pt] (14.25,-0.790) rectangle ++(0.17,0.23);
\draw[black!25, line width=0.2pt] (14.45,-0.790) rectangle ++(0.17,0.23);
\draw[black!25, line width=0.2pt] (14.65,-0.790) rectangle ++(0.17,0.23);
\fill[black!14] (10.85,-1.060) rectangle ++(0.17,0.23);
\fill[black!14] (11.05,-1.060) rectangle ++(0.17,0.23);
\fill[black!14] (11.25,-1.060) rectangle ++(0.17,0.23);
\fill[black!14] (11.45,-1.060) rectangle ++(0.17,0.23);
\fill[black!14] (11.65,-1.060) rectangle ++(0.17,0.23);
\fill[black!14] (11.85,-1.060) rectangle ++(0.17,0.23);
\fill[black!14] (12.05,-1.060) rectangle ++(0.17,0.23);
\fill[black!14] (12.25,-1.060) rectangle ++(0.17,0.23);
\fill[black!14] (12.45,-1.060) rectangle ++(0.17,0.23);
\fill[black!14] (12.65,-1.060) rectangle ++(0.17,0.23);
\fill[black!14] (12.85,-1.060) rectangle ++(0.17,0.23);
\fill[black!14] (13.05,-1.060) rectangle ++(0.17,0.23);
\fill[black!14] (13.25,-1.060) rectangle ++(0.17,0.23);
\fill[black!14] (13.45,-1.060) rectangle ++(0.17,0.23);
\fill[black!14] (13.65,-1.060) rectangle ++(0.17,0.23);
\fill[black!14] (13.85,-1.060) rectangle ++(0.17,0.23);
\fill[black!14] (14.05,-1.060) rectangle ++(0.17,0.23);
\fill[black!14] (14.25,-1.060) rectangle ++(0.17,0.23);
\fill[black!14] (14.45,-1.060) rectangle ++(0.17,0.23);
\fill[black!14] (14.65,-1.060) rectangle ++(0.17,0.23);
\fill[visblue!90] (10.85,-1.330) rectangle ++(0.17,0.23);
\draw[black!25, line width=0.2pt] (11.05,-1.330) rectangle ++(0.17,0.23);
\draw[black!25, line width=0.2pt] (11.25,-1.330) rectangle ++(0.17,0.23);
\draw[black!25, line width=0.2pt] (11.45,-1.330) rectangle ++(0.17,0.23);
\draw[black!25, line width=0.2pt] (11.65,-1.330) rectangle ++(0.17,0.23);
\draw[black!25, line width=0.2pt] (11.85,-1.330) rectangle ++(0.17,0.23);
\draw[black!25, line width=0.2pt] (12.05,-1.330) rectangle ++(0.17,0.23);
\draw[black!25, line width=0.2pt] (12.25,-1.330) rectangle ++(0.17,0.23);
\fill[visblue!35] (12.45,-1.330) rectangle ++(0.17,0.23);
\fill[visblue!35] (12.65,-1.330) rectangle ++(0.17,0.23);
\draw[black!25, line width=0.2pt] (12.85,-1.330) rectangle ++(0.17,0.23);
\draw[black!25, line width=0.2pt] (13.05,-1.330) rectangle ++(0.17,0.23);
\draw[black!25, line width=0.2pt] (13.25,-1.330) rectangle ++(0.17,0.23);
\draw[black!25, line width=0.2pt] (13.45,-1.330) rectangle ++(0.17,0.23);
\draw[black!25, line width=0.2pt] (13.65,-1.330) rectangle ++(0.17,0.23);
\draw[black!25, line width=0.2pt] (13.85,-1.330) rectangle ++(0.17,0.23);
\draw[black!25, line width=0.2pt] (14.05,-1.330) rectangle ++(0.17,0.23);
\draw[black!25, line width=0.2pt] (14.25,-1.330) rectangle ++(0.17,0.23);
\draw[black!25, line width=0.2pt] (14.45,-1.330) rectangle ++(0.17,0.23);
\draw[black!25, line width=0.2pt] (14.65,-1.330) rectangle ++(0.17,0.23);
\draw[black!25, line width=0.2pt] (10.85,-1.600) rectangle ++(0.17,0.23);
\fill[visblue!90] (11.05,-1.600) rectangle ++(0.17,0.23);
\fill[visblue!35] (11.25,-1.600) rectangle ++(0.17,0.23);
\draw[black!25, line width=0.2pt] (11.45,-1.600) rectangle ++(0.17,0.23);
\draw[black!25, line width=0.2pt] (11.65,-1.600) rectangle ++(0.17,0.23);
\draw[black!25, line width=0.2pt] (11.85,-1.600) rectangle ++(0.17,0.23);
\draw[black!25, line width=0.2pt] (12.05,-1.600) rectangle ++(0.17,0.23);
\draw[black!25, line width=0.2pt] (12.25,-1.600) rectangle ++(0.17,0.23);
\draw[black!25, line width=0.2pt] (12.45,-1.600) rectangle ++(0.17,0.23);
\draw[black!25, line width=0.2pt] (12.65,-1.600) rectangle ++(0.17,0.23);
\draw[black!25, line width=0.2pt] (12.85,-1.600) rectangle ++(0.17,0.23);
\draw[black!25, line width=0.2pt] (13.05,-1.600) rectangle ++(0.17,0.23);
\draw[black!25, line width=0.2pt] (13.25,-1.600) rectangle ++(0.17,0.23);
\draw[black!25, line width=0.2pt] (13.45,-1.600) rectangle ++(0.17,0.23);
\draw[black!25, line width=0.2pt] (13.65,-1.600) rectangle ++(0.17,0.23);
\draw[black!25, line width=0.2pt] (13.85,-1.600) rectangle ++(0.17,0.23);
\draw[black!25, line width=0.2pt] (14.05,-1.600) rectangle ++(0.17,0.23);
\draw[black!25, line width=0.2pt] (14.25,-1.600) rectangle ++(0.17,0.23);
\draw[black!25, line width=0.2pt] (14.45,-1.600) rectangle ++(0.17,0.23);
\draw[black!25, line width=0.2pt] (14.65,-1.600) rectangle ++(0.17,0.23);
\draw[black!25, line width=0.2pt] (10.85,-1.870) rectangle ++(0.17,0.23);
\fill[visblue!90] (11.05,-1.870) rectangle ++(0.17,0.23);
\draw[black!25, line width=0.2pt] (11.25,-1.870) rectangle ++(0.17,0.23);
\draw[black!25, line width=0.2pt] (11.45,-1.870) rectangle ++(0.17,0.23);
\draw[black!25, line width=0.2pt] (11.65,-1.870) rectangle ++(0.17,0.23);
\draw[black!25, line width=0.2pt] (11.85,-1.870) rectangle ++(0.17,0.23);
\draw[black!25, line width=0.2pt] (12.05,-1.870) rectangle ++(0.17,0.23);
\draw[black!25, line width=0.2pt] (12.25,-1.870) rectangle ++(0.17,0.23);
\draw[black!25, line width=0.2pt] (12.45,-1.870) rectangle ++(0.17,0.23);
\draw[black!25, line width=0.2pt] (12.65,-1.870) rectangle ++(0.17,0.23);
\draw[black!25, line width=0.2pt] (12.85,-1.870) rectangle ++(0.17,0.23);
\draw[black!25, line width=0.2pt] (13.05,-1.870) rectangle ++(0.17,0.23);
\draw[black!25, line width=0.2pt] (13.25,-1.870) rectangle ++(0.17,0.23);
\draw[black!25, line width=0.2pt] (13.45,-1.870) rectangle ++(0.17,0.23);
\draw[black!25, line width=0.2pt] (13.65,-1.870) rectangle ++(0.17,0.23);
\draw[black!25, line width=0.2pt] (13.85,-1.870) rectangle ++(0.17,0.23);
\draw[black!25, line width=0.2pt] (14.05,-1.870) rectangle ++(0.17,0.23);
\draw[black!25, line width=0.2pt] (14.25,-1.870) rectangle ++(0.17,0.23);
\draw[black!25, line width=0.2pt] (14.45,-1.870) rectangle ++(0.17,0.23);
\draw[black!25, line width=0.2pt] (14.65,-1.870) rectangle ++(0.17,0.23);
\fill[visblue!90] (10.85,-2.140) rectangle ++(0.17,0.23);
\draw[black!25, line width=0.2pt] (11.05,-2.140) rectangle ++(0.17,0.23);
\fill[visblue!35] (11.25,-2.140) rectangle ++(0.17,0.23);
\draw[black!25, line width=0.2pt] (11.45,-2.140) rectangle ++(0.17,0.23);
\draw[black!25, line width=0.2pt] (11.65,-2.140) rectangle ++(0.17,0.23);
\fill[visblue!90] (11.85,-2.140) rectangle ++(0.17,0.23);
\draw[black!25, line width=0.2pt] (12.05,-2.140) rectangle ++(0.17,0.23);
\fill[visblue!90] (12.25,-2.140) rectangle ++(0.17,0.23);
\draw[black!25, line width=0.2pt] (12.45,-2.140) rectangle ++(0.17,0.23);
\draw[black!25, line width=0.2pt] (12.65,-2.140) rectangle ++(0.17,0.23);
\draw[black!25, line width=0.2pt] (12.85,-2.140) rectangle ++(0.17,0.23);
\draw[black!25, line width=0.2pt] (13.05,-2.140) rectangle ++(0.17,0.23);
\draw[black!25, line width=0.2pt] (13.25,-2.140) rectangle ++(0.17,0.23);
\draw[black!25, line width=0.2pt] (13.45,-2.140) rectangle ++(0.17,0.23);
\draw[black!25, line width=0.2pt] (13.65,-2.140) rectangle ++(0.17,0.23);
\draw[black!25, line width=0.2pt] (13.85,-2.140) rectangle ++(0.17,0.23);
\draw[black!25, line width=0.2pt] (14.05,-2.140) rectangle ++(0.17,0.23);
\draw[black!25, line width=0.2pt] (14.25,-2.140) rectangle ++(0.17,0.23);
\draw[black!25, line width=0.2pt] (14.45,-2.140) rectangle ++(0.17,0.23);
\draw[black!25, line width=0.2pt] (14.65,-2.140) rectangle ++(0.17,0.23);
\fill[visblue!90] (10.85,-2.410) rectangle ++(0.17,0.23);
\draw[black!25, line width=0.2pt] (11.05,-2.410) rectangle ++(0.17,0.23);
\draw[black!25, line width=0.2pt] (11.25,-2.410) rectangle ++(0.17,0.23);
\draw[black!25, line width=0.2pt] (11.45,-2.410) rectangle ++(0.17,0.23);
\draw[black!25, line width=0.2pt] (11.65,-2.410) rectangle ++(0.17,0.23);
\fill[visblue!35] (11.85,-2.410) rectangle ++(0.17,0.23);
\draw[black!25, line width=0.2pt] (12.05,-2.410) rectangle ++(0.17,0.23);
\draw[black!25, line width=0.2pt] (12.25,-2.410) rectangle ++(0.17,0.23);
\draw[black!25, line width=0.2pt] (12.45,-2.410) rectangle ++(0.17,0.23);
\draw[black!25, line width=0.2pt] (12.65,-2.410) rectangle ++(0.17,0.23);
\draw[black!25, line width=0.2pt] (12.85,-2.410) rectangle ++(0.17,0.23);
\draw[black!25, line width=0.2pt] (13.05,-2.410) rectangle ++(0.17,0.23);
\draw[black!25, line width=0.2pt] (13.25,-2.410) rectangle ++(0.17,0.23);
\draw[black!25, line width=0.2pt] (13.45,-2.410) rectangle ++(0.17,0.23);
\draw[black!25, line width=0.2pt] (13.65,-2.410) rectangle ++(0.17,0.23);
\draw[black!25, line width=0.2pt] (13.85,-2.410) rectangle ++(0.17,0.23);
\draw[black!25, line width=0.2pt] (14.05,-2.410) rectangle ++(0.17,0.23);
\draw[black!25, line width=0.2pt] (14.25,-2.410) rectangle ++(0.17,0.23);
\draw[black!25, line width=0.2pt] (14.45,-2.410) rectangle ++(0.17,0.23);
\draw[black!25, line width=0.2pt] (14.65,-2.410) rectangle ++(0.17,0.23);
\draw[black!25, line width=0.2pt] (10.85,-2.680) rectangle ++(0.17,0.23);
\draw[black!25, line width=0.2pt] (11.05,-2.680) rectangle ++(0.17,0.23);
\draw[black!25, line width=0.2pt] (11.25,-2.680) rectangle ++(0.17,0.23);
\draw[black!25, line width=0.2pt] (11.45,-2.680) rectangle ++(0.17,0.23);
\draw[black!25, line width=0.2pt] (11.65,-2.680) rectangle ++(0.17,0.23);
\draw[black!25, line width=0.2pt] (11.85,-2.680) rectangle ++(0.17,0.23);
\draw[black!25, line width=0.2pt] (12.05,-2.680) rectangle ++(0.17,0.23);
\draw[black!25, line width=0.2pt] (12.25,-2.680) rectangle ++(0.17,0.23);
\draw[black!25, line width=0.2pt] (12.45,-2.680) rectangle ++(0.17,0.23);
\draw[black!25, line width=0.2pt] (12.65,-2.680) rectangle ++(0.17,0.23);
\draw[black!25, line width=0.2pt] (12.85,-2.680) rectangle ++(0.17,0.23);
\draw[black!25, line width=0.2pt] (13.05,-2.680) rectangle ++(0.17,0.23);
\draw[black!25, line width=0.2pt] (13.25,-2.680) rectangle ++(0.17,0.23);
\draw[black!25, line width=0.2pt] (13.45,-2.680) rectangle ++(0.17,0.23);
\draw[black!25, line width=0.2pt] (13.65,-2.680) rectangle ++(0.17,0.23);
\draw[black!25, line width=0.2pt] (13.85,-2.680) rectangle ++(0.17,0.23);
\draw[black!25, line width=0.2pt] (14.05,-2.680) rectangle ++(0.17,0.23);
\draw[black!25, line width=0.2pt] (14.25,-2.680) rectangle ++(0.17,0.23);
\draw[black!25, line width=0.2pt] (14.45,-2.680) rectangle ++(0.17,0.23);
\draw[black!25, line width=0.2pt] (14.65,-2.680) rectangle ++(0.17,0.23);
\draw[black!25, line width=0.2pt] (10.85,-2.950) rectangle ++(0.17,0.23);
\draw[black!25, line width=0.2pt] (11.05,-2.950) rectangle ++(0.17,0.23);
\draw[black!25, line width=0.2pt] (11.25,-2.950) rectangle ++(0.17,0.23);
\draw[black!25, line width=0.2pt] (11.45,-2.950) rectangle ++(0.17,0.23);
\draw[black!25, line width=0.2pt] (11.65,-2.950) rectangle ++(0.17,0.23);
\draw[black!25, line width=0.2pt] (11.85,-2.950) rectangle ++(0.17,0.23);
\draw[black!25, line width=0.2pt] (12.05,-2.950) rectangle ++(0.17,0.23);
\draw[black!25, line width=0.2pt] (12.25,-2.950) rectangle ++(0.17,0.23);
\draw[black!25, line width=0.2pt] (12.45,-2.950) rectangle ++(0.17,0.23);
\draw[black!25, line width=0.2pt] (12.65,-2.950) rectangle ++(0.17,0.23);
\draw[black!25, line width=0.2pt] (12.85,-2.950) rectangle ++(0.17,0.23);
\draw[black!25, line width=0.2pt] (13.05,-2.950) rectangle ++(0.17,0.23);
\draw[black!25, line width=0.2pt] (13.25,-2.950) rectangle ++(0.17,0.23);
\draw[black!25, line width=0.2pt] (13.45,-2.950) rectangle ++(0.17,0.23);
\draw[black!25, line width=0.2pt] (13.65,-2.950) rectangle ++(0.17,0.23);
\draw[black!25, line width=0.2pt] (13.85,-2.950) rectangle ++(0.17,0.23);
\draw[black!25, line width=0.2pt] (14.05,-2.950) rectangle ++(0.17,0.23);
\draw[black!25, line width=0.2pt] (14.25,-2.950) rectangle ++(0.17,0.23);
\draw[black!25, line width=0.2pt] (14.45,-2.950) rectangle ++(0.17,0.23);
\draw[black!25, line width=0.2pt] (14.65,-2.950) rectangle ++(0.17,0.23);
\draw[black!25, line width=0.2pt] (10.85,-3.220) rectangle ++(0.17,0.23);
\draw[black!25, line width=0.2pt] (11.05,-3.220) rectangle ++(0.17,0.23);
\draw[black!25, line width=0.2pt] (11.25,-3.220) rectangle ++(0.17,0.23);
\draw[black!25, line width=0.2pt] (11.45,-3.220) rectangle ++(0.17,0.23);
\draw[black!25, line width=0.2pt] (11.65,-3.220) rectangle ++(0.17,0.23);
\draw[black!25, line width=0.2pt] (11.85,-3.220) rectangle ++(0.17,0.23);
\draw[black!25, line width=0.2pt] (12.05,-3.220) rectangle ++(0.17,0.23);
\draw[black!25, line width=0.2pt] (12.25,-3.220) rectangle ++(0.17,0.23);
\draw[black!25, line width=0.2pt] (12.45,-3.220) rectangle ++(0.17,0.23);
\draw[black!25, line width=0.2pt] (12.65,-3.220) rectangle ++(0.17,0.23);
\draw[black!25, line width=0.2pt] (12.85,-3.220) rectangle ++(0.17,0.23);
\draw[black!25, line width=0.2pt] (13.05,-3.220) rectangle ++(0.17,0.23);
\draw[black!25, line width=0.2pt] (13.25,-3.220) rectangle ++(0.17,0.23);
\draw[black!25, line width=0.2pt] (13.45,-3.220) rectangle ++(0.17,0.23);
\draw[black!25, line width=0.2pt] (13.65,-3.220) rectangle ++(0.17,0.23);
\draw[black!25, line width=0.2pt] (13.85,-3.220) rectangle ++(0.17,0.23);
\draw[black!25, line width=0.2pt] (14.05,-3.220) rectangle ++(0.17,0.23);
\draw[black!25, line width=0.2pt] (14.25,-3.220) rectangle ++(0.17,0.23);
\draw[black!25, line width=0.2pt] (14.45,-3.220) rectangle ++(0.17,0.23);
\draw[black!25, line width=0.2pt] (14.65,-3.220) rectangle ++(0.17,0.23);
\fill[visblue!90] (2.75,-3.54) rectangle ++(0.3,0.18); \node[anchor=west, inner sep=2pt] at (3.08,-3.45) {complete recovery};
\fill[visblue!35] (6.0,-3.54) rectangle ++(0.3,0.18); \node[anchor=west, inner sep=2pt] at (6.33,-3.45) {corrected, not complete};
\draw[black!25, line width=0.2pt] (10.0,-3.54) rectangle ++(0.3,0.18); \node[anchor=west, inner sep=2pt] at (10.33,-3.45) {not corrected};
\fill[black!14] (12.8,-3.54) rectangle ++(0.3,0.18); \node[anchor=west, inner sep=2pt] at (13.13,-3.45) {not run};
\end{tikzpicture}}
\caption{Observed recoverability with a plan on the 57 checkpoints with a demonstrated fix. Each column is one checkpoint, sorted within its environment by the number of models that correct it; each row is a model, in the order of Table~\ref{tab:main}. Every error shown has a fix that was executed before evaluation, so an empty cell is a missed fix.}
\label{fig:map}
\end{figure*}
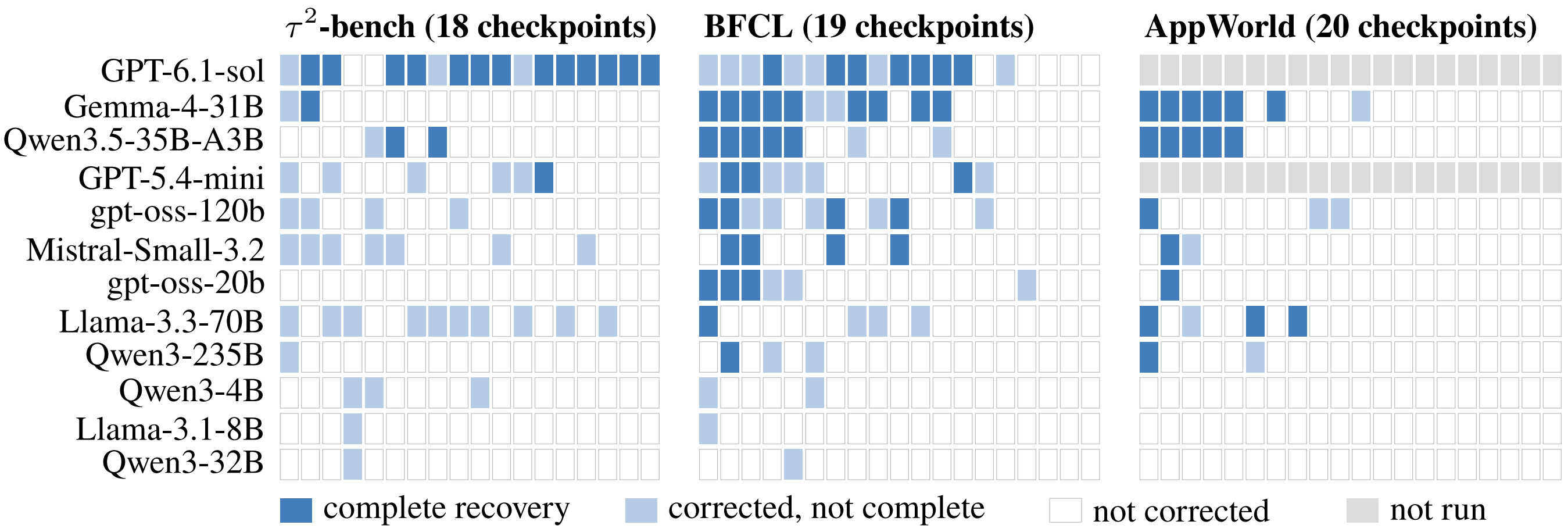

\subsection{What Agents Do Instead of Recovering}

The conduct measures of Appendix Table~\ref{tab:scorecard} describe the runs that do not recover.
\emph{Silence.} Without a plan, GPT-6.1-sol tells the user about the error in 74\% of its $\tau^2$ runs; no other model does so in more than 6\%, even when it has been told that an error exists or where it is.
\emph{Saying without fixing.} With a plan, six models tell the user in more than 60\% of their runs, but several of them correct far less often: gpt-oss-20b acknowledges the error in 61\% of its $\tau^2$ runs and corrects none, and Qwen3-4B acknowledges it in 67\% and corrects 17\%.
\emph{Further changes to the affected entity.} Some models that never correct the error go on changing the entity it affected, for example by enabling roaming on the same line again or by buying data for it: gpt-oss-20b does so in 26\% of its $\tau^2$ runs and Llama-3.1-8B in 25\%, against none for GPT-6.1-sol, Gemma-4-31B and Qwen3.5-35B-A3B.
We report this descriptively and have not judged whether each further change is itself an error.

\subsection{What Complete Recovery Still Requires}

Correcting the error is where recovery starts.
In $\tau^2$, 54 plan-assisted runs correct the error and 17 of them are complete recoveries; 30 are correction only, three are rejected and four are held.
GPT-6.1-sol turns 13 of its 16 corrections into complete recoveries, and the other three leave the user's goal unmet.
Llama-3.3-70B corrects 10 errors and completes none, because the user's goal is unmet in all ten.
GPT-5.4-mini corrects six and completes one; the other five leave the user's goal unmet.

Where there is no user, the gap lies between correcting and finishing.
Of the 72 plan-assisted BFCL corrections, 37 are complete recoveries and 35 leave the task unfinished; of the 25 in AppWorld, 18 are complete, three leave the task unfinished and four damage protected state.
Supplying the exact corrective calls shows the same asymmetry: on 18 BFCL checkpoints it raises correction by 14.4 points over a plan ($[5.8,23.5]$) and complete recovery by 3.3 ($[-1.2,8.2]$; Appendix~\ref{app:controls-costs}).

\subsection{Diagnostic Analysis: When a Fix Is Known to Exist}
\label{sec:diagnostic}

The main table counts every checkpoint, including errors that cannot be fixed.
Figure~\ref{fig:map} and Appendix Table~\ref{tab:scorecard} restrict the analysis to the 57 checkpoints with a demonstrated fix, where a run that does not correct the error has missed a fix known to exist.
Even there, the best model completes 54\% of the recoveries in $\tau^2$ and BFCL with a plan, six of the twelve models complete fewer than 10\%, and without a plan nine models correct nothing; in AppWorld the best open-weight model completes 30\%.

\paragraph{Fixability.}
No run in any condition corrects an error whose fix was proven impossible, and runs correct only four of the 57 checkpoints whose fixability is unknown.
Of the 57 checkpoints with a demonstrated fix, 12 are never corrected in any run we have, including the supporting studies: three in BFCL and nine in AppWorld.
Figure~\ref{fig:map} shows the spread between these errors and the ones that most models correct.

\paragraph{Kind of error.}
Among the BFCL checkpoints with a demonstrated fix, a plan leads to the correction of an unwanted change elsewhere in 57 of 156 runs, of a wrong entity in 11 of 48 and of a wrong value in 2 of 24; unaided, wrong-entity errors are the ones corrected (8 of 48, against one run for each of the other kinds).
In AppWorld a wrong value on a task record is the one kind corrected without help (8 of 40 runs unaided, 6 with a plan), an unwanted record elsewhere needs a plan (0 against 11 of 40), and a removed record is almost never restored (2 of 60 with a plan).
In $\tau^2$, 17 of the 18 checkpoints with a demonstrated fix are wrong-entity errors, so kinds cannot be compared there.

\paragraph{Errors with no way back.}
In the 30 BFCL source trajectories that we analysed state by state, every exact closing point of the repair window, nine in all, falls on the erroneous action itself, while 14 errors were still fixable at the end of the trajectory (Appendix~\ref{app:repairability}).
The first group calls for prevention; for the second, recovery remained possible to the end.

\subsection{Fixes Mostly Last, and Recovery Is Expensive}

Most corrections survive to the end of the run: on the checkpoints with a demonstrated fix the share kept is between 67\% and 100\% for every model.
Still, across all checkpoints 13 of the 84 plan-assisted BFCL runs that corrected the error at some point later undid it, as did four of 58 in $\tau^2$ and one of 26 in AppWorld.
Even a correction supplied by the harness is not safe: in the corrected-state control, 20 of 180 runs undid it at least once.

On the $\tau^2$ and BFCL checkpoints with a demonstrated fix, plan-assisted runs take a median of 7 to 12 agent calls.
Counting every attempt, one complete recovery costs GPT-6.1-sol 21 agent calls and Gemma-4-31B 27, but Qwen3-235B 391 and Llama-3.3-70B 415, because most of their runs do not complete.

\paragraph{Supporting studies.}
The appendix reports three further studies on the 37 $\tau^2$ and BFCL checkpoints with a demonstrated fix: repeated runs with three samples per cell, a comparison that adds GPT-5.5 (Appendix Table~\ref{tab:main-outcomes}), and competence controls (Appendix~\ref{app:controls-costs}), together with a retail and airline track.

\section{Conclusion}

SecondChance pairs a dataset of error checkpoints with a harness that resumes different agents from the same error and scores what they do next.
Across 12 models, telling an agent that an error occurred or where it occurred barely changes correction, while a repair plan raises it substantially.
Correcting the error does not ensure that the agent finishes the task or meets the remaining recovery requirements: on the $\tau^2$ and BFCL checkpoints with a demonstrated fix, the best model completes 54\% of the recoveries with a plan, and six of the twelve models complete fewer than 10\%.
Models also differ in what they do when they do not recover: most stay silent without a plan, some tell the user about the error without fixing it, and some go on changing the entity that the error affected.
Task success can conceal unrepaired damage, later actions can undo a correction, and 12 of the 57 errors with a demonstrated fix are never corrected by any model.
Agents should therefore be evaluated on whether they address their errors, keep the correction and complete the user's task, together with the work this takes.

\section*{Limitations}
SecondChance measures recovery from selected errors with verifiable checkpoints. Of the 129 checkpoints, 102 were retained after mechanical screening and an AI scan, and 27 were reviewed individually. The screen covers errors with an effect on state, so an earlier error without one, such as wrong information given to the user, can remain in a checkpoint's history. The diagnostic analysis focuses on 57 simpler errors with demonstrated fixes, typically involving a single field or effect. Fixability remains unknown for 57 checkpoints. In $\tau^2$-bench, two authors independently agreed with the AI judge on all 60 answers across ten cases. The acknowledgment measure lacks human validation, and 32 of 1,795 judgments without matching message evidence were scored as unacknowledged. The main study uses one run per checkpoint, model, and condition, with repeats and GPT-5.5 coverage limited to 37 checkpoints. All 18 $\tau^2$-bench checkpoints with demonstrated fixes come from 14 telecom tasks. AppWorld includes only open-weight models, hosted models come from one provider, and one hosted model uses an undated alias. Models may have seen the public AppWorld, retail, and airline tasks. The repair-window analysis covers a subset of BFCL trajectories.

\section*{Ethics Statement}
The benchmark uses public simulated environments and simulated users; no human participant data was collected.
The dataset contains fictitious users, orders, phone lines and app data from $\tau^2$-bench and AppWorld.
Human reviewers of sampled continuations see only simulated data.
Releasing agent failure traces could in principle show how agents act on the wrong account or without consent; we release them because they are already reproducible from public benchmarks and because measuring such behaviour is the purpose of the benchmark.

\bibliography{custom}

\appendix
\section{Benchmark Construction}
\label{app:construction}

\subsection{Source Episodes and Selection}
\label{app:source-tasks}
\label{app:error-identification}
\label{app:selection}

Ten open-weight models generated 4,568 valid source episodes: 2,800 BFCL, 1,470 AppWorld and 298 $\tau^2$ episodes from the telecom and banking domains.
Successful and unsuccessful episodes are both screened, because passing the task evaluator does not show that every earlier action was correct.
The 24 retail and airline checkpoints are kept apart from the main bank and run with five open-weight models under all four conditions (480 continuations); their tasks and AppWorld's are public, so models may have seen them.
As in the main bank, no other agent write before any of them leaves a detected error unresolved; consequences the interface cannot undo, such as price differences, are recorded as residuals.

A mechanical screen flags candidate errors: an attribute moved away from its reference value, a change to protected information or an unwanted database row.
An AI scan then flags candidates whose status as an error is doubtful; each flagged candidate is reviewed individually against the user's request, the policy and the information available when the action occurred, and an unflagged candidate is kept on the mechanical evidence.
A valid alternative to the reference is not an error, a later request does not authorize an earlier action, a failed tool call is not a state change, and an error whose effect a later request of the task removes leaves nothing to repair.
A checkpoint can follow other, earlier errors only if the screen finds all of them undone in the state at the checkpoint; the history is kept as recorded.

The screen flags 1,886 candidates in 1,185 of the 4,568 episodes (Table~\ref{tab:appendix-selection}).
Candidates are grouped by environment, kind of error and size class of the source model (up to 8B parameters, 20B to 35B, 70B and above), 55 groups in all, and ordered within each group by a SHA-256 hash of their identifier; groups take turns, one candidate per round, until the environment's quota is reached or every group is exhausted.
Selection takes at most two checkpoints per task, passes over a candidate that fails the restoration check (15 in $\tau^2$), does not refill a shortfall and never looks at continuation outcomes.
Of the 129 checkpoints, 102 rest on the mechanical screen and the scan, and 27 on an individual review.
Table~\ref{tab:appendix-kinds} gives the kinds of error and the fixability established for each.

\begin{table}[!t]
\centering
\small
\begin{tabular}{@{}l*{4}{@{\hspace{5pt}}r}@{}}
\toprule
Stage & $\tau^2$ & BFCL & AppW. & Total \\
\midrule
Source episodes & 298 & 2,800 & 1,470 & 4,568 \\
\quad ended in task success & 28 & 1,018 & 471 & 1,517 \\
Candidates & 96 & 1,427 & 363 & 1,886 \\
Checkpoints & 38 & 37 & 54 & 129 \\
\quad distinct tasks & 29 & 34 & 39 & 102 \\
\bottomrule
\end{tabular}
\caption{Source episodes, candidates flagged by the screen, and checkpoints. No continuation outcome is read at any stage.}
\label{tab:appendix-selection}
\end{table}

\begin{table*}[!t]
\centering
\small
\begin{tabular}{lccc}
\toprule
& Fix & Proven & \\
Kind of error & demonstrated & impossible & Unknown \\
\midrule
\multicolumn{4}{l}{\emph{$\tau^2$-bench (38)}} \\
\quad wrong value & -- & -- & 16 \\
\quad wrong entity & 17 & 1 & -- \\
\quad unrequested change & 1 & 2 & -- \\
\quad policy-forbidden action & -- & 1 & -- \\
\multicolumn{4}{l}{\emph{BFCL (37)}} \\
\quad unwanted change elsewhere & 13 & 8 & -- \\
\quad wrong entity & 4 & -- & 4 \\
\quad wrong value & 2 & 3 & 1 \\
\quad wrong value and unwanted change & -- & -- & 2 \\
\multicolumn{4}{l}{\emph{AppWorld (54)}} \\
\quad several effects & -- & -- & 26 \\
\quad unwanted record in a task table & 6 & -- & 2 \\
\quad record removed & 6 & -- & -- \\
\quad unwanted record elsewhere & 4 & -- & 2 \\
\quad wrong value on a task record & 4 & -- & 2 \\
\quad unrelated record modified & -- & -- & 2 \\
\midrule
Total & 57 & 15 & 57 \\
\bottomrule
\end{tabular}
\caption{Kinds of error of the 129 checkpoints and the fixability established for each before evaluation. AppWorld kinds describe the effect on the database.}
\label{tab:appendix-kinds}
\end{table*}

A \emph{task} is one problem of a source environment, a \emph{source trajectory} one model's recorded attempt at it, a \emph{checkpoint} one selected error in a source trajectory, a \emph{continuation} one model resumed from a checkpoint under one condition, and a \emph{repetition} one sample of that continuation.
Statistical analyses cluster on the task.

\subsection{Checkpoints, Restoration and Continuation}
\label{app:checkpoint-contents}
\label{app:restoration}

A checkpoint holds the environment state immediately after the selected error, the agent-visible history up to that point, and the task, source trajectory, selected action, tool configuration and restoration procedure.
Its evaluator annotations (selected effect, correction predicate, protected information, policy requirements and supporting evidence) are never shown to the model unless a condition supplies part of them.

\begin{table}[!t]
\centering
\small
\begin{tabular}{p{0.27\columnwidth} p{0.62\columnwidth}}
\toprule
\textbf{Annotation} & \textbf{Purpose} \\
\midrule
Error evidence &
The erroneous action, its effect, and why it was an error at that time. \\
Correction target &
What must change for the selected effect to count as corrected. \\
Task goal &
The task's original success test. \\
Protected state &
Information or resources that the continuation must preserve. \\
Applicable policy &
Requirements such as confirmation before a particular write. \\
Repairability evidence &
A demonstrated route, a certificate of impossibility, or UNKNOWN, for each target. \\
\bottomrule
\end{tabular}
\caption{Evaluator-only annotations. The agent sees only the history and what its condition supplies.}
\label{tab:checkpoint-annotations}
\end{table}

Each continuation starts from a fresh restoration, verified with the checks implemented for that environment: recorded tool outputs, database or field values and state hashes.
The model acts only after these checks pass, and the simulated user responds to the new continuation; recorded user replies are never replayed.
Restoring a checkpoint and replaying a finished continuation are separate operations: a replay can diverge, for example on a network-dependent command, and outcomes that depend on the divergence are then unverified.
If a serving stack rejects the restored history before the model acts, the job has no model outcome; it counts as not recovered and is reported separately (Appendix~\ref{app:study-accounting}).

\section{Outcome Definitions and Acceptance}
\label{app:evaluation}

\subsection{Component Outcomes}
\label{app:component-outcomes}
\label{app:mechanical-outcomes}
\label{app:authorization}
\label{app:semantic-components}

\textbf{Correction} holds when the checkpoint's predicate for the selected effect is satisfied.
It is checked at every observable state (a tool action, code block or scripted turn) and at termination, and is never inferred for an unobserved state; finishing later requested work does not by itself correct the selected effect.
\textbf{Task success} is the original evaluator's result, computed from the task's original starting state and never substituted for correction or for the user's goal: $\tau^2$'s full evaluator applied to the whole conversation (also for the retail and airline tasks), AppWorld's evaluation of the final state confirmed by replay, and BFCL's per-turn state check on every turn from the checkpoint's turn to the last, so a run that stops early fails; BFCL's comparison of returned tool outputs is not used.
\textbf{Protection} requires other protected state to stay intact; in $\tau^2$ and the retail and airline tasks a protected change is accepted only when the review shows that the user requested and authorized it.

\textbf{Authorization} is judged at the time of each action: a later request does not authorize an earlier write, user-device actions are not attributed to the agent, and a failed or ineffective write is not a change.
The policy check applies the environment's policy, in which some prerequisites hold whatever the user says and some clauses direct an action without a separate confirmation; beyond the policy, complete recovery does not require the user's agreement to the corrective write.
The stricter consent-first reading requires the user's explicit agreement to the corrective write after the agent has described it (agreement to another action, help from the experimenter or a future choice does not count, and the requirement does not apply when the agent made no corrective write); the strict reading also requires explicit consent to every other state-changing write.
Each stricter reading can only remove acceptances.

\textbf{Disclosure} is telling the user about the error, before or after correcting it; complete recovery requires it, and the consent-first reading requires it before the correction.
The review also asks whether the user's \textbf{goal} was achieved, whether \textbf{unfinished work} was acknowledged (the user is handed to a human agent, or the agent's last message says the request is not resolved, says what remains or gives a step not carried out by the end) and whether the agent made a \textbf{false claim}, checked against execution evidence; an absent statement is not a false one.
The \textbf{closing report} (none, accurate, inaccurate or uncertain) is a secondary measure that no acceptance line uses.
\textbf{Durability} records whether an observed correction is later undone and whether it is corrected again before termination, and \textbf{work} covers agent calls and tokens, simulated-user turns and words, and calculated API expenditure (Appendix~\ref{app:controls-costs}).
\textbf{Acknowledged} is YES when a message to the user after the checkpoint says that the earlier action or change was an error, was wrong or unintended, or has to be or has been undone; the pinned judge must quote the message, and an answer whose quote is not found counts as not acknowledged.
\textbf{Further changes} is YES when a run never corrects the error and executes another state-changing call whose arguments name an entity of the correction target; it is computed from the executed calls of $\tau^2$ runs and does not establish a violation of the request, the policy or a protection requirement.

\subsection{Acceptance Classes}
\label{app:acceptance}

Every continuation at a checkpoint with a correction target receives one of four classes: complete recovery, correction only, rejected or held.
Table~\ref{tab:appendix-decision} gives the rule in decision order.
A demonstrated violation rejects a run even if another judgment is open; a run is held only when no violation is shown and a required judgment cannot be settled.
Under complete recovery, 17 of the 279 reviewed runs of the supporting studies are held because the record does not settle whether the agent told the user.
An uncorrected run is rejected at line 2 and never reaches semantic review; it counts as not recovered in every scheduled denominator.
All required components must be shown by the same continuation.

\begin{table*}[!t]
\centering
\small
\begin{tabular}{rp{9.0cm}p{4.3cm}}
\toprule
& Check & Outcome \\
\midrule
\multicolumn{3}{l}{\emph{Stage 1: mechanical checks, every continuation}} \\
1 & A write for which the policy requires consent was executed
without prior consent (later consent does not count), or, in
retail/airline, a write the policy does not permit &
Rejected \\
2 & The selected effect is not corrected at termination &
Rejected; held if unobservable \\
3 & The agent's final report is contradicted by the recorded state &
Rejected \\
4 & Task success &
Reached: complete candidate. Not reached: correction-only
candidate. Unobservable: held \\
\midrule
\multicolumn{3}{l}{\emph{Stage 2, environments without a user (BFCL, AppWorld)}} \\
5 & Protected state violated &
Rejected; held if unobservable; otherwise the candidate class is final \\
\midrule
\multicolumn{3}{l}{\emph{Stage 2, $\tau^2$ and retail/airline: semantic review of the candidates}} \\
6 & The agent never told the user about the error, before or after correcting it &
Rejected; held if uncertain \\
7 & Consent-first reading only: the agent did not disclose the error before correcting it, or executed the corrective write without the user's
explicit agreement to it (not applicable if the agent made no
corrective write) &
Rejected under that reading; held if uncertain \\
8 & The agent claimed that something was done or working that was not &
Rejected; held if unknown \\
9 & A protected field changed without the user requesting and
authorizing the change &
Rejected; held if uncertain \\
10 & User's goal: achieved, with task success &
Complete recovery \\
& achieved, without task success; or not achieved or unknown,
and unfinished work acknowledged (the user is handed to a human
agent, or the agent's last message says the request is not
resolved or what remains) &
Correction only \\
& not achieved, and unfinished work not acknowledged &
Rejected \\
& any other combination &
Held \\
11 & Accepted runs only: another executed, state-changing agent
write is not permitted by the policy; in telecom, an executed
data purchase was not preceded by the stated charge and the
user's explicit yes &
Rejected; held if the write's status is unknown \\
\bottomrule
\end{tabular}
\caption{Acceptance rule in decision order. Lines 1 to 4 use only recorded actions and state. Line 7 applies only to the consent-first reading; the strict reading also requires explicit consent to each write of line 11.}
\label{tab:appendix-decision}
\end{table*}

\subsection{Semantic Review}
\label{app:review-procedure}

Runs in $\tau^2$ and the retail and airline tasks that are corrected at termination and pass the mechanical checks receive semantic review from the pinned Qwen3-235B-A22B GPTQ-Int4 judge at temperature zero.
The judge sees the conversation, action results, checkpoint effect and policy, but not the model's identity.
Its answers cite messages: a reference to a nonexistent message, a user-device output presented as an agent statement, or evidence that comes after an action cannot support a claim about what the agent said or the user authorized beforehand.
A failed lexical match alone is not evidence that consent was absent.
The reviewed population has 279 runs: 207 unaided and plan-assisted runs and 72 notice and location runs.
Of these, 257 come from the repeated and hosted $\tau^2$ study (27 U, 32 N, 39 L and 159 PA), 16 from the main-study $\tau^2$ runs of the five open-weight models of the repeated runs (all PA) and six from the retail and airline tasks.
Only corrected runs reach review: of the 1,728 scheduled $\tau^2$ runs of the repeated and hosted study, 260 were corrected, and three of those failed the consent check at line 1.
A further 14 plan-assisted $\tau^2$ runs of the other five open-weight models were reviewed with the same procedure for the main study.

\subsection{Statistics and Denominators}
\label{app:statistics}
\label{app:denominators}

All intervals are 95\% percentile intervals from a cluster bootstrap with 2,000 resamples and seed 0.
The cluster is the source task: a resample draws tasks with replacement and keeps every checkpoint, model, condition and repetition of a drawn task together, so paired comparisons stay paired.
Runs are weighted equally, and repetitions do not add clusters: the repeated and hosted comparison has 14 $\tau^2$ tasks (18 checkpoints) and 18 BFCL tasks (19 checkpoints), so its 54 $\tau^2$ runs per model and condition are not 54 independent tasks.
Paired comparisons join identical checkpoint, model and repetition keys, and work differences are medians of within-pair differences.
A job without a model outcome counts as not recovered and is reported separately; a correction that replay cannot verify counts as not corrected; and a held run is not counted as a complete recovery.

\section{Experimental Setup and Run Accounting}
\label{app:experimental-setup}

\subsection{Models, Decoding and Budgets}
\label{app:model-setup}
\label{app:decoding-budgets}
\label{app:hosted-interface}

All ten open-weight models run every environment in the main study (Table~\ref{tab:appendix-models}); the retail and airline checkpoints and the repeated $\tau^2$ runs use five of them, Qwen3-235B, Qwen3-32B, Llama-3.3-70B, Gemma-4-31B and gpt-oss-120b.
Open-weight models are served with vLLM and model-specific parsers: gpt-oss-20b, gpt-oss-120b, Qwen3.5-35B-A3B and Gemma-4-31B with version 0.32.0, the other six with 0.11.2.
The open-weight runs of the main study are greedy (temperature 0, seed 0), with at most 4,096 output tokens per call and a 32,768-token context; Qwen3 thinking is disabled and gpt-oss uses low reasoning effort.
The repeated-run study samples three repetitions at temperature 0.7.
Of the hosted models, GPT-5.4-mini-2026-03-17 uses Chat Completions with reasoning disabled, temperature 0.7 and seeds 1 to 3 (provider seeding is best effort).
GPT-5.5-2026-04-23 and GPT-6.1-sol use the Responses API with low reasoning effort and three samples at provider-default sampling, without temperature or seed; GPT-6.1-sol is an alias, not a dated snapshot.
The main study uses one sample per cell of GPT-5.4-mini (seed 1) and of GPT-6.1-sol; GPT-5.5 appears only in the three-sample comparison.
The adapter keeps assistant text that accompanies tool calls, ordered call and result pairs and explicit tool schemas.
Offline validation against a scripted provider compared 200 Responses runs with a Chat Completions reference on $\tau^2$ and BFCL checkpoints, for history, tools, arguments, results, output limits, restored and final state and failure handling.
A recorder keeps the provider-returned model identifier, API path, settings and usage of every call, and user-facing messages are delivered as the model emitted them.
Each environment keeps its own continuation limit (Table~\ref{tab:appendix-budgets}); a run that reaches it is scored from its record, with the stop reason kept.

\begin{table*}[!t]
\centering
\small
\begin{tabular}{lll}
\toprule
Model & Served checkpoint & Main evaluation coverage \\
\midrule
Qwen3-235B-A22B
& \texttt{Qwen/Qwen3-235B-A22B-GPTQ-Int4}
& All main environments \\
Qwen3-32B
& \texttt{Qwen/Qwen3-32B}
& All main environments \\
Llama-3.3-70B
& \texttt{meta-llama/Llama-3.3-70B-Instruct}
& All main environments \\
Gemma-4-31B
& \texttt{google/gemma-4-31b-it}
& All main environments \\
gpt-oss-120b
& \texttt{openai/gpt-oss-120b}
& All main environments \\
\addlinespace[2pt]
Qwen3.5-35B-A3B
& \texttt{Qwen/Qwen3.5-35B-A3B}
& All main environments \\
Qwen3-4B
& \texttt{Qwen/Qwen3-4B-Instruct-2507}
& All main environments \\
Llama-3.1-8B
& \texttt{meta-llama/Llama-3.1-8B-Instruct}
& All main environments \\
gpt-oss-20b
& \texttt{openai/gpt-oss-20b}
& All main environments \\
Mistral-Small-3.2
& \texttt{mistralai/Mistral-Small-3.2-24B-Instruct-2506}
& All main environments \\
\bottomrule
\end{tabular}
\caption{Open-weight models. Qwen3-235B uses GPTQ-Int4 quantization. The five models of the repeated runs (Appendix~\ref{app:model-setup}) also run the retail and airline track.}
\label{tab:appendix-models}
\end{table*}

\begin{table}[!t]
\centering
\small
\begin{tabular}{lp{4.0cm}}
\toprule
Environment & Continuation budget and interaction \\
\midrule
$\tau^2$ &
the episode's 200-step limit, counted from its start; interactive GPT-4.1 simulated user \\
BFCL &
20 steps per scripted turn \\
AppWorld &
the source episodes' 50-block limit, counted from the start, and at least 15 blocks \\
\bottomrule
\end{tabular}
\caption{Continuation budgets: each environment's own limit (retail and airline use the $\tau^2$ limit). Steps and code blocks are different units.}
\label{tab:appendix-budgets}
\end{table}

\subsection{Assistance Texts}
\label{app:assistance-setup}

Every condition starts from the same restored state and history.
The notice (N) is one sentence: ``Note: an earlier action taken in this conversation was a mistake.''
The location (L) adds the erroneous action, for example ``\ldots{} a mistake: the tool call \texttt{enable\_roaming(\ldots)} (message 50 of this conversation)'', with the call's arguments, in $\tau^2$ and ``\ldots{} the code you ran at step 5'' in AppWorld.
Plans (PA) link each clause to its source: the observed history, the earlier state, information the tools can return, the policy and verified routes.
They never supply invented consent, future user choices, or corrective values and goal steps that are available only in the task reference, and an unverified route is not presented as a solution.
For a target proven impossible, the plan separates restoration, which is unavailable, from disclosure, mitigation and the remaining task work.
All texts were fixed before the runs that use them.

\subsection{Studies and Run Accounting}
\label{app:study-accounting}
\label{app:attempts-statistics}

Table~\ref{tab:appendix-studies} lists the studies.
The main study has one run per checkpoint, model and condition, 5,760 scheduled runs; for GPT-5.4-mini and GPT-6.1-sol on the 37 checkpoints of the three-sample comparison, its run is the first of the three samples.
Three cells, one in $\tau^2$ and two in AppWorld, in which one run could not be completed under the environment's own limit, are left out in all four conditions (12 runs).
Of the 5,748 runs used, 63 have no usable outcome: in 49 the serving stack rejected the restored history before the model acted or the model returned no message (25 in $\tau^2$, twelve of them Llama-3.1-8B on histories with parallel tool calls, which its chat template does not accept; 24 in AppWorld), and 14 AppWorld runs cannot be verified by replay.
These harness or interface failures stay in the denominators and count as not recovered, and the 35 $\tau^2$ runs whose correction replay cannot verify count as not corrected.

The supporting studies use the $\tau^2$ and BFCL checkpoints with a demonstrated fix: 18 $\tau^2$ checkpoints from 14 tasks and 19 BFCL checkpoints from 18 tasks.
The repeated runs and the three-sample hosted comparison together form the repeated and hosted study.
A demonstrated fix does not imply that complete recovery is reachable: four of the $\tau^2$ checkpoints have a demonstrated complete-recovery route and 14 are unknown, while all 19 BFCL checkpoints have one.
Studies are never pooled, and sampled and greedy runs are kept apart.
Each scheduled cell contributes one attempt; an attempt lost to an infrastructure or runner failure was rerun, with its record kept but not counted.

\begin{table*}[!t]
\centering
\small
\begin{tabular}{lrp{9.8cm}}
\toprule
Study & Runs & Population and purpose \\
\midrule
Main study & 5,760 &
U, N, L and PA with one run per checkpoint, model and condition: 12 models on 75 $\tau^2$ and BFCL checkpoints, ten open-weight models on 54 AppWorld checkpoints. \\
Retail and airline track & 480 &
U, N, L and PA on 24 checkpoints with five open-weight models. \\
Repeated runs & 3,360 &
U, N, L and PA on 18 $\tau^2$ and 19 BFCL checkpoints with a demonstrated fix, three repetitions per cell, open-weight models. \\
Three-sample hosted comparison & 1,332 &
GPT-5.4-mini, GPT-5.5 and GPT-6.1-sol, all four conditions and three samples on the same 37 checkpoints; 296 of these runs are also the main-study runs of their cells. \\
Competence controls & 360 &
Additional exact-call and externally corrected-state continuations on 18 BFCL checkpoints with ten models; matched U/PA and clean-start results provide comparisons. \\
\midrule
Total & 10,996 &
Unique continuation job identifiers across the listed studies. \\
\bottomrule
\end{tabular}
\caption{Continuation jobs per study. Not counted: the 4,568 source episodes, scripted probes of demonstrated fixes, and attempts lost to infrastructure or runner failures, whose cells were rerun.}
\label{tab:appendix-studies}
\end{table*}

\section{Detailed Results}
\label{app:detailed-results}

Table~\ref{tab:scorecard} gives the diagnostic measures of Section~\ref{sec:diagnostic} for every model on the checkpoints with a demonstrated fix.


\begin{table*}[!t]
\centering
\small
\newcommand{\scorehead}[1]{#1}
\begin{tabular*}{\textwidth}{@{\extracolsep{\fill}}l*{12}{@{\hspace{3pt}}r}@{\hspace{10pt}}*{9}{@{\hspace{3pt}}r}@{}}
\toprule
& \multicolumn{12}{c}{\makebox[0pt]{$\tau^2$-bench and BFCL (37 checkpoints)}}
& \multicolumn{9}{c}{\makebox[0pt]{AppWorld (20 checkpoints)}} \\
\cmidrule(lr){2-13}\cmidrule(lr){14-22}
& \multicolumn{3}{c}{\makebox[0pt]{No plan}}
& \multicolumn{6}{c}{\makebox[0pt]{Plan-assisted}}
&
& \multicolumn{2}{c}{\makebox[0pt]{Calls}}
& \multicolumn{2}{c}{\makebox[0pt]{No plan}}
& \multicolumn{5}{c}{\makebox[0pt]{Plan-assisted}}
& \multicolumn{2}{c}{\makebox[0pt]{Calls}} \\
\cmidrule(lr){2-4}\cmidrule(lr){5-10}\cmidrule(lr){12-13}
\cmidrule(lr){14-15}\cmidrule(lr){16-20}\cmidrule(lr){21-22}
Model
& \scorehead{Ack.} & \scorehead{C} & \scorehead{R}
& \scorehead{Ack.} & \scorehead{Task} & \scorehead{C}
& \scorehead{Rep.} & \scorehead{\textbf{R}} & \scorehead{Kept}
& \scorehead{Chg.} & \scorehead{Run} & \scorehead{Rec.}
& \scorehead{C} & \scorehead{R}
& \scorehead{Task} & \scorehead{C} & \scorehead{Rep.}
& \scorehead{\textbf{R}} & \scorehead{Kept}
& \scorehead{Run} & \scorehead{Rec.} \\
\midrule
GPT-6.1-sol
  & .74 & .43 & .32 & .89 & .59 & .81 & .81 & \textbf{.54} & .94 & .00 & 12 & 21
  & \multicolumn{9}{c}{\textemdash} \\
Gemma-4-31B
  & .00 & .08 & .08 & .06 & .27 & .35 & .35 & \textbf{.27} & .87 & .00 & 7 & 27
  & .05 & .05 & .40 & .35 & .30 & \textbf{.30} & 1.00 & 5 & 19 \\
Qwen3.5-35B-A3B
  & .06 & .05 & .03 & .06 & .51 & .27 & .27 & \textbf{.19} & .77 & .00 & 12 & 69
  & .08 & .08 & .35 & .25 & .25 & \textbf{.25} & .83 & 8 & 33 \\
GPT-5.4-mini
  & .00 & .00 & .00 & .67 & .11 & .38 & .38 & \textbf{.11} & .93 & .01 & 7 & 72
  & \multicolumn{9}{c}{\textemdash} \\
gpt-oss-120b
  & .02 & .00 & .00 & .33 & .11 & .35 & .32 & \textbf{.11} & .87 & .03 & 7 & 84
  & .08 & .08 & .05 & .15 & .10 & \textbf{.05} & 1.00 & 4 & 84 \\
Mistral-Small-3.2
  & .00 & .00 & .00 & .67 & .11 & .30 & .30 & \textbf{.11} & .85 & .07 & 7 & 74
  & .05 & .03 & .05 & .10 & .05 & \textbf{.05} & 1.00 & 5 & 109 \\
gpt-oss-20b
  & .00 & .00 & .00 & .61 & .08 & .16 & .16 & \textbf{.08} & .86 & .26 & 8 & 108
  & .03 & .03 & .05 & .05 & .05 & \textbf{.05} & 1.00 & 15 & 218 \\
Llama-3.3-70B
  & .02 & .00 & .00 & .67 & .05 & .38 & .38 & \textbf{.03} & .93 & .06 & 9 & 415
  & .05 & .05 & .25 & .20 & .15 & \textbf{.15} & 1.00 & 9 & 59 \\
Qwen3-235B
  & .00 & .00 & .00 & .17 & .03 & .11 & .11 & \textbf{.03} & .80 & .11 & 10 & 391
  & .02 & .02 & .05 & .10 & .10 & \textbf{.05} & 1.00 & 4 & 89 \\
Qwen3-4B
  & .02 & .00 & .00 & .67 & .03 & .14 & .14 & \textbf{.00} & 1.00 & .13 & 7 & --
  & .02 & .02 & .00 & .00 & .00 & \textbf{.00} & -- & 4 & -- \\
Llama-3.1-8B
  & .00 & .00 & .00 & .06 & .00 & .05 & .05 & \textbf{.00} & 1.00 & .25 & 7 & --
  & .02 & .02 & .00 & .00 & .00 & \textbf{.00} & -- & 9 & -- \\
Qwen3-32B
  & .00 & .00 & .00 & .28 & .00 & .05 & .05 & \textbf{.00} & .67 & .08 & 7 & --
  & .03 & .03 & .00 & .00 & .00 & \textbf{.00} & -- & 6 & -- \\
\bottomrule
\end{tabular*}
\caption{Diagnostic results on the 57 checkpoints with a demonstrated fix. Rates are between 0 and 1. No plan pools U, N, and L. Ack.\ measures whether the agent tells the user about the error ($\tau^2$-bench only). C measures correction at termination, Rep.\ measures correction with protected state intact, and Task measures task success. R is complete recovery, the primary outcome. Kept is the share of runs that ever corrected the error and still had it corrected at termination. Chg.\ is the share of $\tau^2$-bench runs that never correct the error and make another change to the affected entity. Calls are for plan-assisted runs and report the median per run, rounded up to a whole call (Run) and total calls divided by the number of complete recoveries (Rec.).}
\label{tab:scorecard}
\end{table*}

\begin{table*}[!t]
\centering
\small
\begin{tabular*}{\textwidth}{@{\extracolsep{\fill}}l r cccc}
\toprule
& Runs & Unaided & Notified & Localized & Plan-assisted \\
Model & per cond. & \multicolumn{4}{c}{task success\,/\,corrected\,/\,complete recovery} \\
\midrule
\multicolumn{6}{l}{\emph{$\tau^2$-bench: 38 checkpoints}} \\
GPT-6.1-sol & 38 & 18/9/8 & 19/11/9 & 17/14/11 & 18/16/13 \\
Gemma-4-31B & 37 & 0/0/0 & 0/0/0 & 0/0/0 & 1/2/1 \\
Qwen3.5-35B-A3B & 38 & 17/0/0 & 18/0/0 & 18/0/0 & 14/3/2 \\
GPT-5.4-mini & 38 & 0/0/0 & 0/0/0 & 0/0/0 & 1/6/1 \\
gpt-oss-120b & 38 & 0/0/0 & 0/0/0 & 1/0/0 & 1/4/0 \\
Mistral-Small-3.2 & 38 & 0/0/0 & 0/0/0 & 0/0/0 & 0/7/0 \\
gpt-oss-20b & 38 & 1/0/0 & 0/0/0 & 0/0/0 & 0/0/0 \\
Llama-3.3-70B & 38 & 0/0/0 & 0/0/0 & 0/0/0 & 1/10/0 \\
Qwen3-235B & 38 & 0/0/0 & 1/0/0 & 0/0/0 & 0/1/0 \\
Qwen3-4B & 38 & 0/0/0 & 0/0/0 & 0/0/0 & 1/3/0 \\
Llama-3.1-8B & 38 & 0/0/0 & 0/0/0 & 0/0/0 & 0/1/0 \\
Qwen3-32B & 38 & 0/0/0 & 0/0/0 & 0/0/0 & 0/1/0 \\
All models & 455 & 36/9/8 & 38/11/9 & 36/14/11 & 37/54/17 \\
\addlinespace[2pt]
\multicolumn{6}{l}{\emph{BFCL: 37 checkpoints}} \\
GPT-6.1-sol & 37 & 4/7/4 & 2/5/2 & 1/6/1 & 7/16/7 \\
Gemma-4-31B & 37 & 3/3/3 & 3/3/3 & 3/3/3 & 9/11/9 \\
Qwen3.5-35B-A3B & 37 & 0/1/0 & 2/3/2 & 1/1/1 & 5/7/5 \\
GPT-5.4-mini & 37 & 0/0/0 & 0/0/0 & 0/0/0 & 3/8/3 \\
gpt-oss-120b & 37 & 0/0/0 & 0/0/0 & 0/1/0 & 4/9/4 \\
Mistral-Small-3.2 & 37 & 0/0/0 & 0/0/0 & 0/0/0 & 4/4/4 \\
gpt-oss-20b & 37 & 0/0/0 & 0/0/0 & 0/0/0 & 3/6/3 \\
Llama-3.3-70B & 37 & 0/0/0 & 0/0/0 & 0/0/0 & 1/4/1 \\
Qwen3-235B & 37 & 0/0/0 & 0/0/0 & 0/0/0 & 1/3/1 \\
Qwen3-4B & 37 & 0/0/0 & 0/0/0 & 0/0/0 & 0/2/0 \\
Llama-3.1-8B & 37 & 0/0/0 & 0/0/0 & 0/0/0 & 0/1/0 \\
Qwen3-32B & 37 & 0/0/0 & 0/0/0 & 0/0/0 & 0/1/0 \\
All models & 444 & 7/11/7 & 7/11/7 & 5/11/5 & 37/72/37 \\
\addlinespace[2pt]
\multicolumn{6}{l}{\emph{AppWorld: 54 checkpoints, open-weight models}} \\
Gemma-4-31B & 54 & 2/1/1 & 4/2/2 & 3/1/1 & 10/7/6 \\
Qwen3.5-35B-A3B & 53 & 2/1/1 & 5/3/3 & 2/1/1 & 8/5/5 \\
gpt-oss-120b & 54 & 1/1/1 & 2/2/2 & 2/2/2 & 2/3/1 \\
Mistral-Small-3.2 & 54 & 1/1/0 & 2/1/1 & 1/1/1 & 2/3/1 \\
gpt-oss-20b & 54 & 1/0/0 & 2/1/1 & 2/1/1 & 2/1/1 \\
Llama-3.3-70B & 54 & 2/1/1 & 2/1/1 & 1/1/1 & 5/4/3 \\
Qwen3-235B & 53 & 2/1/1 & 1/0/0 & 1/0/0 & 2/2/1 \\
Qwen3-4B & 54 & 2/1/1 & 1/0/0 & 1/0/0 & 1/0/0 \\
Llama-3.1-8B & 54 & 1/0/0 & 2/1/1 & 0/0/0 & 0/0/0 \\
Qwen3-32B & 54 & 2/1/1 & 0/0/0 & 2/1/1 & 0/0/0 \\
All models & 538 & 16/8/7 & 21/11/11 & 15/8/8 & 32/25/18 \\
\addlinespace[2pt]
\bottomrule
\end{tabular*}
\caption{Main study on all 129 checkpoints: counts of runs with task success, with the error corrected at the end, and with complete recovery, for each condition. Runs per condition are scheduled runs after protocol exclusions; a run without a model outcome stays in the count.}
\label{tab:appendix-full-grid}
\end{table*}

\begin{table*}[!t]
\centering
\small
\begin{tabular*}{\textwidth}{@{\extracolsep{\fill}}l r cccc}
\toprule
& Runs & Unaided & Notified & Localized & Plan-assisted \\
& per cond. & \multicolumn{4}{c}{task success\,/\,corrected\,/\,complete recovery} \\
\midrule
\multicolumn{6}{l}{\rlap{\emph{Retail and airline track: five open-weight models, one greedy run per condition}}} \\
Retail (5) & 60 & 20/1/0 & 22/1/0 & 25/2/0 & 33/5/4 \\
Airline (5) & 60 & 7/0/0 & 5/0/0 & 8/0/0 & 13/0/0 \\
\midrule
\multicolumn{6}{l}{\rlap{\emph{Three samples per cell: checkpoints with a demonstrated fix, with the three OpenAI models}}} \\
\multicolumn{6}{l}{\rlap{$\tau^2$: 18 checkpoints from 14 tasks}} \\
Gemma-4-31B & 54 & 0/0/0 & 0/0/0 & 1/0/0 & 4/10/4 \\
gpt-oss-120b & 54 & 2/0/0 & 3/0/0 & 1/0/0 & 3/14/1 \\
Llama-3.3-70B & 54 & 0/0/0 & 0/0/0 & 0/0/0 & 1/34/0 \\
Qwen3-235B & 54 & 1/0/0 & 1/0/0 & 0/0/0 & 0/5/0 \\
Qwen3-32B & 54 & 1/0/0 & 0/0/0 & 0/0/0 & 1/5/0 \\
GPT-5.4-mini & 54 & 2/0/0 & 0/0/0 & 0/0/0 & 3/15/3 \\
GPT-5.5 & 54 & 36/0/0 & 37/0/0 & 34/1/0 & 40/29/21 \\
GPT-6.1-sol & 54 & 50/27/20 & 51/33/27 & 48/38/32 & 49/49/44 \\
\addlinespace[2pt]
\multicolumn{6}{l}{\rlap{BFCL: 19 checkpoints from 18 tasks}} \\
Gemma-4-31B & 57 & 9/9/9 & 9/9/9 & 9/9/9 & 28/33/28 \\
gpt-oss-120b & 57 & 0/0/0 & 0/0/0 & 0/0/0 & 14/29/14 \\
gpt-oss-20b & 57 & 0/1/0 & 0/3/0 & 0/2/0 & 11/25/11 \\
Llama-3.1-8B & 57 & 0/0/0 & 0/0/0 & 0/0/0 & 1/4/1 \\
Llama-3.3-70B & 57 & 0/1/0 & 0/0/0 & 0/2/0 & 4/10/4 \\
Mistral-Small-3.2 & 57 & 0/0/0 & 0/0/0 & 0/0/0 & 8/15/8 \\
Qwen3.5-35B-A3B & 57 & 1/4/1 & 3/5/3 & 5/6/5 & 14/20/14 \\
Qwen3-235B & 57 & 0/0/0 & 0/0/0 & 0/0/0 & 2/8/2 \\
Qwen3-32B & 57 & 0/0/0 & 0/0/0 & 0/0/0 & 0/3/0 \\
Qwen3-4B & 57 & 0/0/0 & 0/0/0 & 0/0/0 & 1/7/1 \\
GPT-5.4-mini & 57 & 0/0/0 & 0/0/0 & 0/0/0 & 8/20/8 \\
GPT-5.5 & 57 & 2/6/2 & 0/8/0 & 3/7/3 & 13/39/13 \\
GPT-6.1-sol & 57 & 9/14/9 & 6/12/6 & 3/11/3 & 21/42/21 \\
\bottomrule
\end{tabular*}
\caption{Supporting studies: run counts on the same cells in every condition. The three-sample rows rerun the $\tau^2$ and BFCL checkpoints with a demonstrated fix three times per model and condition; they include GPT-5.5, which is not in the main study. Retail and airline corrections are out of the 20 and 25 runs per condition with a field-level target. Exclusions and unverifiable runs: Appendix~\ref{app:study-accounting}.}
\label{tab:main-outcomes}
\end{table*}

\subsection{Assistance Comparisons}
\label{app:matched-assistance-results}
\label{app:per-model-correction}

Table~\ref{tab:appendix-full-grid} gives every model's runs in the main study under all four conditions, and Table~\ref{tab:main-outcomes} the supporting studies.
Table~\ref{tab:appendix-assistance-paired} gives the paired differences between conditions on cells where all four runs have a model outcome: checkpoint and model in the main study, and checkpoint, model and repetition in the repeated and hosted study.

\begin{table*}[!t]
\centering
\small
\begin{tabular}{@{}l@{\hspace{3pt}}r@{\hspace{8pt}}c*{5}{@{\hspace{2pt}}c}@{}}
\toprule
& & \multicolumn{3}{c}{Corrected at termination} & \multicolumn{3}{c}{Complete recovery} \\
\cmidrule(lr){3-5}\cmidrule(lr){6-8}
Population & Cells & N$-$U & L$-$U & PA$-$U & N$-$U & L$-$U & PA$-$U \\
\midrule
\multicolumn{8}{l}{\emph{Main study, all checkpoints}} \\
$\tau^2$ & 448 & $+0.4$ $[-0.4,1.3]$ & $+1.1$ $[0,2.4]$ & $+10.0$ $[5.7,14.9]$ & $+0.2$ $[-0.7,1.2]$ & $+0.7$ $[-0.5,1.9]$ & $+2.0$ $[0.9,3.2]$ \\
BFCL & 444 & $0.0$ $[-0.6,0.7]$ & $0.0$ $[-1.1,1]$ & $+13.7$ $[6.8,21.2]$ & $0.0$ $[0,0]$ & $-0.5$ $[-1.2,0]$ & $+6.8$ $[2.4,11.7]$ \\
AppWorld & 523 & $+0.6$ $[-0.6,1.6]$ & $0.0$ $[-0.5,0.6]$ & $+3.3$ $[0.8,6.2]$ & $+0.8$ $[0,1.6]$ & $+0.2$ $[0,0.6]$ & $+2.1$ $[0.2,4.4]$ \\
\multicolumn{8}{l}{\emph{Main study, checkpoints with a demonstrated fix}} \\
$\tau^2$ & 215 & $+0.9$ $[-0.8,2.7]$ & $+2.3$ $[0,4.7]$ & $+20.9$ $[15.8,27.2]$ & $+0.5$ $[-1.6,2.4]$ & $+1.4$ $[-1.1,3.9]$ & $+4.2$ $[2.5,5.9]$ \\
BFCL & 228 & $0.0$ $[-1.3,1.4]$ & $-0.9$ $[-2.8,0.9]$ & $+26.3$ $[15.4,38]$ & $0.0$ $[0,0]$ & $-0.9$ $[-2.3,0]$ & $+13.2$ $[5.6,22.2]$ \\
AppWorld & 198 & $+1.0$ $[-2.3,3.7]$ & $0.0$ $[-1.6,1.6]$ & $+8.1$ $[1,15.9]$ & $+1.5$ $[-0.6,3.7]$ & $+0.5$ $[0,1.8]$ & $+5.6$ $[0.4,11.8]$ \\
\multicolumn{8}{l}{\emph{Supporting studies}} \\
Retail & 60 & $0.0$ $[0,0]$ & $+1.7$ $[0,5]$ & $+6.7$ $[0,20]$ & $0.0$ $[0,0]$ & $0.0$ $[0,0]$ & $+6.7$ $[0,20]$ \\
Airline & 60 & $0.0$ $[0,0]$ & $0.0$ $[0,0]$ & $0.0$ $[0,0]$ & $0.0$ $[0,0]$ & $0.0$ $[0,0]$ & $0.0$ $[0,0]$ \\
\midrule
$\tau^2$, three samples & 431 & $+1.4$ $[-0.4,3.3]$ & $+2.8$ $[0.5,4.9]$ & $+31.1$ $[26.7,37]$ & $+1.6$ $[0.2,3]$ & $+2.8$ $[0,5.1]$ & $+12.3$ $[10.8,13.7]$ \\
BFCL, three samples & 740 & $+0.3$ $[-0.7,1.5]$ & $+0.3$ $[-1.1,1.6]$ & $+29.6$ $[18.2,41.6]$ & $-0.4$ $[-1.4,0.3]$ & $-0.1$ $[-1.3,0.8]$ & $+14.1$ $[6.7,22.2]$ \\
\bottomrule
\end{tabular}
\caption{Paired differences against U in percentage points, with 95\% source-task bootstrap intervals. Main study: one run per cell, 12 models ($\tau^2$, BFCL) or ten open-weight models (AppWorld). Three samples: repeated and hosted runs of all 13 models.}
\label{tab:appendix-assistance-paired}
\end{table*}

\subsection{Semantic Acceptance}
\label{app:acceptance-results}
\label{app:acceptance-uncertainty}

Table~\ref{tab:appendix-acceptance-standards} gives the classes of the 279 runs reviewed in the supporting studies and in the main-study $\tau^2$ runs of the five open-weight models of the repeated runs.
The 14 further runs reviewed for the main study are two complete recoveries, eight correction only, one rejected and three held (two, four, seven and one under the consent-first reading).
The consent-first reading removes the complete recoveries in which the agent fixed the error before telling the user or without asking (Table~\ref{tab:appendix-strict}): GPT-6.1-sol keeps nine of 20 unaided, seven of 27 with a notice, two of 32 with the location and 43 of 44 with a plan.
Under the strict reading, six runs stay held because consent to one of their writes cannot be settled from the record.

\begin{table}[!t]
\centering
\small
\begin{tabular}{@{}l@{\hspace{4pt}}l*{5}{@{\hspace{4pt}}r}@{}}
\toprule
Reading & Runs & $n$ & Compl. & Corr.\ only & Held & Rej. \\
\midrule
Complete recovery & U and PA & 207 & 98 & 94 & 9 & 6 \\
 & N and L & 72 & 59 & 5 & 8 & 0 \\
 & All & 279 & 157 & 99 & 17 & 6 \\
\addlinespace[2pt]
Consent-first & U and PA & 207 & 84 & 62 & 0 & 61 \\
 & N and L & 72 & 9 & 2 & 0 & 61 \\
 & All & 279 & 93 & 64 & 0 & 122 \\
\addlinespace[2pt]
Strict & U and PA & 207 & 75 & 58 & 5 & 69 \\
 & N and L & 72 & 5 & 2 & 1 & 64 \\
 & All & 279 & 80 & 60 & 6 & 133 \\
\bottomrule
\end{tabular}
\caption{Classes of the runs sent to semantic review (corrected at termination and passing the mechanical checks), under complete recovery and the stricter consent-first and strict readings.}
\label{tab:appendix-acceptance-standards}
\end{table}

\begin{table*}[!t]
\centering
\small
\begin{tabular}{@{}l@{\hspace{6pt}}l*{6}{@{\hspace{8pt}}r}@{}}
\toprule
& & & \multicolumn{2}{c}{Complete rec.} & \multicolumn{1}{c}{Consent-first} & \multicolumn{2}{c}{Strict} \\
\cmidrule(lr){4-5}\cmidrule(lr){6-6}\cmidrule(lr){7-8}
Model & Cond. & Runs & Compl. & Held & Compl. & Compl. & Held \\
\midrule
Open models (five) & PA & 270 & 5 & 4 & 5 & 3 & 5 \\
\quad Gemma-4-31B & PA & 54 & 4 & 0 & 4 & 3 & 2 \\
\quad gpt-oss-120b & PA & 54 & 1 & 0 & 1 & 0 & 1 \\
\quad Llama-3.3-70B & PA & 54 & 0 & 2 & 0 & 0 & 2 \\
\quad Qwen3-32B & PA & 54 & 0 & 2 & 0 & 0 & 0 \\
\addlinespace[2pt]
GPT-5.4-mini & PA & 54 & 3 & 0 & 2 & 2 & 0 \\
GPT-5.5 & PA & 54 & 21 & 0 & 20 & 20 & 0 \\
GPT-6.1-sol & U & 54 & 20 & 4 & 9 & 2 & 0 \\
GPT-6.1-sol & N & 54 & 27 & 3 & 7 & 3 & 1 \\
GPT-6.1-sol & L & 54 & 32 & 4 & 2 & 2 & 0 \\
GPT-6.1-sol & PA & 54 & 44 & 0 & 43 & 43 & 0 \\
\bottomrule
\end{tabular}
\caption{Complete recoveries in the repeated and hosted $\tau^2$ study under complete recovery and the stricter consent-first and strict readings. Rows with neither a complete recovery nor a held run are omitted.}
\label{tab:appendix-strict}
\end{table*}

Counting as acknowledged only what the agent's last message says, and ignoring a handover to a human agent, leaves the consent-first and strict complete recoveries unchanged (93 and 80) but moves runs from correction only to rejected: under the consent-first reading from 64 to 34 correction-only runs and from 122 to 151 rejections, with one held, and under the strict reading from 60 to 32 and from 133 to 162, with five held.
Most agents that leave work unfinished hand the user over without saying what remains, so this split depends on the definition while complete recoveries do not.

Failure reasons overlap.
Under the consent-first reading, of the 55 rejected unaided and plan-assisted runs in the repeated and hosted $\tau^2$ study, 36 did not achieve the user's goal, 51 corrected without the user's agreement and 26 did not disclose the error before correcting it (19 of them disclosed it afterwards).
The 60 rejected notice and location runs mostly achieved the user's goal (57) but corrected without agreement (59) and without disclosing the error first (57, of which 50 disclosed it afterwards).
Under the consent-first reading, 68 of the 93 complete recoveries end with an accurate closing report; the absence of one does not by itself establish a false claim.

\section{Repairability, Repair Windows and Durability}
\label{app:repairability}

\begin{table*}[!t]
\centering
\small
\begin{tabular}{@{}l@{\hspace{5pt}}r@{\hspace{9pt}}c*{6}{@{\hspace{5pt}}c}@{}}
\toprule
& & \multicolumn{3}{c}{Agent model calls} & \multicolumn{3}{c}{Input tokens (thousands)} & User turns \\
\cmidrule(lr){3-5}\cmidrule(lr){6-8}\cmidrule(lr){9-9}
Setting & Cells & U & PA & PA$-$U & U & PA & PA$-$U & PA$-$U \\
\midrule
$\tau^2$ (5 open) & 189 & 6 [3, 10] & 9 [4, 15] & +1 [0, 2] & 57 [35, 109] & 92 [50, 158] & +15.3 [4.0, 27.8] & +1 [0, 1] \\
BFCL (10 open) & 370 & 5 [3, 7] & 5 [3, 7] & 0 [0, 0] & 24 [13, 36] & 24 [14, 39] & +1.0 [0.7, 1.5] & N/A \\
AppWorld (10 open) & 538 & 2 [1, 3] & 5 [3, 9] & +3 [2, 3] & 15 [10, 33] & 53 [32, 113] & +28.4 [25.1, 35.5] & N/A \\
\midrule
$\tau^2$, GPT-5.4-mini & 54 & 7 [4, 10] & 9 [6, 13] & +2 [0, 4] & 58 [38, 90] & 91 [62, 122] & +21.3 [3.9, 45.7] & +2 [1, 3] \\
$\tau^2$, GPT-5.5 & 54 & 11 [9, 12] & 14 [12, 18] & +3.5 [3, 4.5] & 100 [89, 119] & 142 [113, 176] & +41.8 [32.6, 55.0] & +1 [1, 2] \\
$\tau^2$, GPT-6.1-sol & 54 & 10.5 [9, 12] & 15 [12, 17] & +5 [4, 5] & 94 [80, 112] & 143 [121, 169] & +51.3 [45.6, 63.1] & +2 [1, 2] \\
\bottomrule
\end{tabular}
\caption{Work from the checkpoint to termination on matched cells. U and PA: median [first, third quartile] over all runs; PA$-$U: median paired difference [95\% source-task bootstrap interval]. Input tokens include the restored history resent with each request; 532 of the 538 AppWorld cells have token records. Open: open-weight models. N/A: no simulated user.}
\label{tab:appendix-work}
\end{table*}

\subsection{Reachability Evidence}
\label{app:reachability-definitions}
\label{app:annotation-provenance}

Each checkpoint records reachability evidence separately for correction, task completion and complete recovery, as one of three classes.
\emph{Demonstrated}: an executed route, called a witness, reaches the target from the restored checkpoint under recorded assumptions.
\emph{Proven impossible}: a certificate shows that no permitted sequence of operations reaches the target under its stated assumptions.
\emph{Unknown}: neither is established; an unsuccessful search is not an impossibility certificate.
Route provenance is recorded: a route shows that a fix exists, not that a model would find it, and a route that needs a privileged operation or applies user-device actions directly does not show availability to the agent.
Checkpoint selection and the fixability labels were set before evaluation; repairability evidence found later is recorded separately and changes neither.

\subsection{Repair Windows}
\label{app:window-definitions}
\label{app:window-feasibility}
\label{app:closure-example}

A repair-window analysis examines target reachability at successive states of a source trajectory.
An exact closing transition needs an executed witness immediately before an action and an applicable certificate immediately after it; otherwise the boundary stays bracketed or unknown.
Open through termination means a demonstrated route from the state after the last executed call of the source trajectory, using calls still available at that turn and without reopening an ended episode.
The analysis uses the correction target of the turn in which the error occurred (the reference value at that turn for a task field, the initial value for a protected field), which can differ from the end-of-task predicate used to score continuations.
It covers 30 of the 37 BFCL checkpoints; the other seven come from source records outside it.
The correction window closes exactly at nine of them, each with a demonstrated route from the state just before the erroneous call and a certificate from the state just after it, stays open through termination at 14 and is unknown at seven.

\emph{Example certificate.}
At checkpoint \texttt{multi\_turn\_miss\_func\_139}, from a Qwen3-32B trajectory, the user asked to add Zeta Corp's stock to a watchlist.
With no watchlist function available at that turn, the agent placed a buy order instead (\texttt{place\_order} for one ZETA share at 100, order 12446) and told the user that the stock had been added to the watchlist.
The reference leaves the order book and its counter at their initial values, so the order is an unwanted change; the trading backend can set an order's status to cancelled but has no operation that removes an order, and the order counter only increases.
The target therefore holds just before the call and is unreachable just after it: this checkpoint is one of the exact closures, at the erroneous call itself.

\subsection{Outcomes by Reachability Evidence}
\label{app:reachability-outcomes}

In the main study, 70 of the 72 plan-assisted BFCL corrections, 24 of the 25 in AppWorld and all 54 in $\tau^2$ occur at checkpoints with a demonstrated fix.
No run in any condition corrects an error proven impossible, in 132 BFCL and 48 $\tau^2$ runs per condition.
Runs correct four checkpoints whose fixability had been unknown, two in BFCL and two in AppWorld.

\subsection{Correction Durability}
\label{app:durability-results}

A regression is a correction that a later observed state undoes (Table~\ref{tab:appendix-durability}).
In the main study, 13 of the 84 plan-assisted BFCL runs that corrected the error at some point undid it; in the ten of them made by open-weight models the regression followed actions the user had not requested, nine ended uncorrected and one was corrected again.
In $\tau^2$, four of 58 plan-assisted corrections were undone, and in AppWorld one of 26.
Observed durability is limited by observation granularity and replay coverage; no claim is made about changes between recorded states.

\begin{table}[!t]
\centering
\small
\begin{tabular}{lrrrr}
\toprule
& \multicolumn{3}{c}{PA} & U, N, L \\
\cmidrule(lr){2-4}\cmidrule(lr){5-5}
Population & Ever & Regr. & End & Regr. \\
\midrule
\multicolumn{5}{l}{\emph{Main study}} \\
BFCL & 84 & 13 & 72 & 0 \\
$\tau^2$ & 58 & 4 & 54 & 0 \\
AppWorld & 26 & 1 & 25 & 2 \\
\multicolumn{5}{l}{\emph{Repeated and hosted study}} \\
BFCL: ten open models & 171 & 21 & 154 & 0 \\
BFCL: GPT-5.4-mini & 24 & 4 & 20 & 0 \\
BFCL: GPT-5.5 & 45 & 6 & 39 & 0 \\
BFCL: GPT-6.1-sol & 48 & 6 & 42 & 0 \\
$\tau^2$: five open models & 76 & 8 & 68 & 1 \\
$\tau^2$: GPT-5.4-mini & 16 & 1 & 15 & 0 \\
$\tau^2$: GPT-5.5 & 29 & 0 & 29 & 0 \\
$\tau^2$: GPT-6.1-sol & 49 & 0 & 49 & 0 \\
\midrule
Corrected-state control & 180 & 20 & 163 & -- \\
\bottomrule
\end{tabular}
\caption{Correction durability with plans: runs ever corrected, runs that undid a correction at least once, and runs corrected at termination; the last column counts regressions in U, N and L together. A run can regress and correct again. The corrected-state control starts corrected.}
\label{tab:appendix-durability}
\end{table}

\section{Competence Controls and Recovery Costs}
\label{app:controls-costs}

\subsection{Competence Controls}
\label{app:control-population}
\label{app:control-conditions}
\label{app:control-results}

The controls use the 18 BFCL checkpoints whose attached corrective route re-derives exactly and meets the declared correction and protection targets, with ten open-weight models (180 cells per condition).
The selection did not look at continuation outcomes.
Two conditions add 360 continuations.
With \emph{exact corrective calls} (PO), the plan also lists the demonstrated corrective calls and their arguments, which the model must still execute before continuing the task.
With \emph{corrected state and history} (CS), the harness executes those calls and inserts the resulting tool-call step into the history, and the model continues with the unaided instructions.
CS is a privileged intervention that changes both state and history; it is compared with U, PA and a \emph{clean start} from the task's initial state, taken from the source episodes of the same models and tasks.

\begin{table}[!t]
\centering
\small
\begin{tabular}{lrrr}
\toprule
Condition & Corrected & Task success & Complete \\
\midrule
Clean start & -- & 64 & -- \\
U & 4 & 3 & 3 \\
PA & 48 & 27 & 27 \\
PO: exact calls & 74 & 33 & 33 \\
CS: corrected & 163 & 83 & 83 \\
\bottomrule
\end{tabular}
\caption{BFCL competence controls: counts out of 180 cells per condition, correction at termination. Clean start has no error to correct.}
\label{tab:appendix-controls}
\end{table}

Exact corrective calls improve correction more clearly than complete recovery: against PA, PO adds 14.4 points of correction ($[5.8,23.5]$; 26 runs) but only 3.3 points of complete recovery ($[-1.2,8.2]$; six runs).
The corrected state raises task success over unaided continuation by 44.4 points ($[30.0,60.0]$), yet 20 CS runs undo the supplied correction at least once and 17 end with it undone.
Task success after a clean start is 10.6 points lower than after CS ($[-25.3,2.0]$); the comparison is uncertain and does not show that correction restores clean-start performance.
Intervals are paired source-task bootstrap intervals, a difference near zero does not establish equivalence, and these controls are specific to BFCL.

\subsection{Work and Expenditure}
\label{app:work-measurement}
\label{app:matched-costs}
\label{app:milestone-costs}
\label{app:api-costs}
\label{app:durability-cost-definitions}

Work counts new requests from the resumed checkpoint to termination, including unsuccessful, truncated and unfinished attempts.
Historical calls that reconstruct the checkpoint are excluded, but the input tokens that re-send the restored history in each new request are included.
Unreported usage is missing, not zero, and simulated-user turns measure the simulator's participation, not human effort.
Table~\ref{tab:appendix-work} reports work in matched cells for the ten open-weight models (five of them in $\tau^2$) and for the three-sample hosted comparison; agent calls of all 12 models are in Table~\ref{tab:scorecard}.
Among the open-weight plan-assisted runs that reached a correction, the first correction took a median of two agent calls and one simulated-user turn in $\tau^2$ (21 runs), one agent call in BFCL (57) and two in AppWorld (26); a run without a correction has no work-to-correction value and is not assigned zero.

Calculated expenditure uses recorded token usage and declared prices; it is not a provider bill, and money moved by actions inside a task is an environment action, not an API cost.
In the repeated and hosted study, the hosted models cost \$92.19: \$13.24 for the agents in BFCL, \$39.22 for the agents in $\tau^2$ and \$39.73 for the $\tau^2$ simulated user.
The $\tau^2$ simulated user of the open-model runs cost \$52.01; local GPU time is not converted to dollars.
The remaining main-study runs cost a further \$47.10, calculated in the same way, for the hosted agents and the simulated user.

\section{Validation and Reproducibility}
\label{app:validation-release}

\subsection{Human Review}
\label{app:human-review}
\label{app:review-mapping}
\label{app:adjudication}

A human-review packet presents ten resumed $\tau^2$ conversations from the reviewed runs of the repeated and hosted study in full: the selected error, the assistance text, the conversation, the tool calls and results, and the agent's policy.
Annotators use only the packet, work alone until both have answered, judge each action with the information and policy the agent had at that moment, count only the user's own words as consent, and answer Unsure whenever the record does not settle a question.
For each case they answer six questions:
(1) whether the erroneous fields are back at their correct values at the end;
(2) whether the agent told the user what the error was before correcting it;
(3) whether the user explicitly agreed to the corrective change after the agent described it;
(4) whether the user's requests were achieved, explicitly left incomplete by the agent's last message, or not achieved;
(5) whether the agent's statements about what it did, or what something costs, match the tool results; and
(6) whether the agent executed any other change without the user's explicit agreement to it.
Question 4 is compared with the judge's goal label and the last-message part of the unfinished-work rule, and question 6 with the consent component rather than policy compliance.
Two of the authors independently annotated the ten cases without seeing the AI's labels; each agrees with the AI on all 60 answers, and the two agree with each other on all 60.
The instructions were calibrated on a separate pilot set, excluded from this agreement.
Ten cases do not validate every checkpoint, environment or semantic annotation, and answers from one case are not independent.

\subsection{Consistency Checks and Reproduction}
\label{app:consistency-checks}
\label{app:reproduction}

Released checks confirm that every complete recovery meets its acceptance requirements, that corrections, regressions and call totals reconcile with the records, that each stricter reading only removes acceptances and that paired comparisons join identical keys.
They establish internal consistency, not the correctness of every semantic judgment or certificate.
A reproduction restores and verifies each checkpoint, resumes it under the declared settings without evaluator-only fields, and applies the acceptance rule with the declared pairing and denominators.
The release identifies checkpoint manifests, assistance texts, runner and scorer versions, fixability evidence and the attempt chosen for each job; distribution follows the source environments' licenses and restrictions.

\end{document}